# CP-nets: A Tool for Representing and Reasoning with Conditional *Ceteris Paribus* Preference Statements

**Craig Boutilier** CEBLY@CS.TORONTO.EDU
*Department of Computer Science*
*University of Toronto*
*Toronto, ON, M5S 3H8, Canada*

**Ronen I. Brafman** BRAFMAN@CS.BGU.AC.IL
*Department of Computer Science*
*Ben-Gurion University*
*Beer Sheva, Israel 84105*

**Carmel Domshlak** DCARMEL@CS.CORNELL.EDU
*Department of Computer Science*
*Cornell University*
*Ithaca, NY 14853, USA*

**Holger H. Hoos** HOOS@CS.UBC.CA
*Department of Computer Science*
*University of British Columbia*
*Vancouver, BC, V6T 1Z4, Canada*

**David Poole** POOLE@CS.UBC.CA
*Department of Computer Science*
*University of British Columbia*
*Vancouver, BC, V6T 1Z4, Canada*

## Abstract

Information about user preferences plays a key role in automated decision making. In many domains it is desirable to assess such preferences in a qualitative rather than quantitative way. In this paper, we propose a qualitative graphical representation of preferences that reflects conditional dependence and independence of preference statements under a *ceteris paribus* (all else being equal) interpretation. Such a representation is often compact and arguably quite natural in many circumstances. We provide a formal semantics for this model, and describe how the structure of the network can be exploited in several inference tasks, such as determining whether one outcome dominates (is preferred to) another, ordering a set outcomes according to the preference relation, and constructing the best outcome subject to available evidence.

## 1. Introduction

Extracting preference information from users is generally an arduous process, and human decision analysts have developed sophisticated techniques to help elicit this information (Howard & Matheson, 1984). A key goal in the study of computer-based decision support is the construction of tools that allow the preference elicitation process to be automated, either partially or fully. Methods for extracting, representing and reasoning about the preferences of naive users are particularly important in AI applications, ranging from collaborative filtering (Lashkari, Metral, & Maes, 1994) and recommender systems (Nguyen & Haddawy, 1998) to product configuration (D'Ambrosio & Birmingham, 1995) and medical decision making (Chajewska, Getoor, Norman, & Shahar, 1998). In many of these applications users cannot be expected to have the patience (or sometimes the ability) to provide detailed preference relations or utility functions. Typical users may not be able to provide much more than qualitative rankings of fairly circumscribed outcomes.

In this paper we describe a novel graphical representation, *CP-nets*, that can be used for specifying preference relations in a relatively compact, intuitive, and structured manner using conditional *ceteris paribus* (all else being equal) preference statements. CP-nets can be used to specify different types of preference relations, such as a preference ordering over potential decision outcomes or a likelihood ordering over possible states of the world, for example, as in Shoham's (1987) preference semantics. However, it is mainly the first type—preferences over the outcomes of decisions—that motivates the development of CP-nets. The inference techniques for CP-nets described in this paper focus on two important, related questions: how to perform preferential comparison between outcomes, and how to find the optimal outcome given a partial assignment to the problem attributes.

Ideally, a preference representation should capture statements that are natural for users to assess, be reasonably compact, and support effective inference. Our conditional *ceteris paribus* semantics requires that the user specify, for any specific attribute $A$ of interest, which other attributes can impact her preferences for values of $A$. For each instantiation of the relevant attributes—the *parents* of $A$—the user must specify her preference ordering over values of $A$ conditional on the parents assuming the instantiated values; for instance, $a_1$ may be preferred to $a_2$ when $b_1$ and $c_2$ hold. Such a preference is given a *ceteris paribus* interpretation: $a_1$ is preferred to $a_2$ given $b_1$ and $c_2$ *all else being equal*. In other words, for any fixed instantiation of the remaining attributes, an outcome where $a_1$ holds is preferred to one where $a_2$ holds (assuming $b_1$ and $c_2$). Such statements are arguably quite natural and appear in several places (e.g., in e-commerce applications). For instance, the product selection service offered by Active Buyer's Guide[1] asks for (unconditional) *ceteris paribus* statements in assessing a user's preference for various products. The tools there also ask for certain semi-quantitative information about preferences. Conditional expressions offer even greater flexibility.

Preference elicitation is a complex task and is a key focus of work in decision analysis (Keeney & Raiffa, 1976; Howard & Matheson, 1984; French, 1986), especially elicitation involving non-expert users. Automating the process of preference extraction can be very difficult. There has been considerable work on exploiting the structure of preferences and utility functions in a way that allows them to be appropriately decomposed (Keeney & Raiffa, 1976; Bacchus & Grove, 1995, 1996; La Mura & Shoham, 1999). For instance, if certain attributes are *preferentially independent* of others (Keeney & Raiffa, 1976), one can assign degrees of preference to these attribute values without worrying about other attribute values. Furthermore, if one assumes more stringent conditions, often one can construct an additive value function in which each attribute contributes to overall preference to a specific "degree" (the *weight* of that attribute) (Keeney & Raiffa, 1976). For instance, it is common in some engineering design problems to make such assumptions and simply

1. See `www.activebuyersguide.com`.

require users to assess the weights (D'Ambrosio & Birmingham, 1995). This allows the direct tradeoffs between values of different attributes to be assessed concisely. Case-based approaches have also recently been considered (Ha & Haddawy, 1998).

Models such as these make the preference elicitation process easier by imposing specific requirements on the form of the utility or preference function. We consider our CP-net representation to offer an appropriate tradeoff between allowing flexible preference expression and imposing a particular preference structure. Specifically, unlike much of the work cited above, CP-nets capture conditional preference statements.

The remainder of the paper is organized as follows. Section 2 provides background on preference orderings, and important notions such as preferential independence and conditional *ceteris paribus* preference statements. We then define CP-nets, discussing their semantics and expressive power in depth, and some of the model's properties. In Section 3 we present an algorithm for outcome optimization in CP-nets and provide an example of an application of CP-nets that illustrates the optimization process. Section 4 introduces two kinds of queries for preferential comparison, namely, ordering and dominance queries, and investigates their computational properties. Section 5 discusses several general techniques for answering dominance queries that exploit the structure of a CP-net. In Section 6 we discuss the applicability of our complexity results and algorithms to a slight generalization of CP-nets that allow incompletely specified local preferences and/or statements of preferential indifference. Finally, in Section 7 we examine some related work and applications of CP-nets, and discuss a number of interesting directions for future theoretical research and applications.

## 2. Model Definition

Philosophical treatment of many intuitive qualitative preferential statements began in 1957 in a pioneering work of Halldén (1957), and was continued by Castañeda (1958), von Wright (1963, 1972), Kron and Milovanović (1975), Trapp (1985), and Hansson (1996). The reason for such an intensive analysis of these statements is expressed concisely in the opening of Hansson's (1996) paper:

> When discussing with my wife what table to buy for our living room, I said: "A round table is better than a square one." By this I did not mean that irrespectively of their other properties, any round table is better than any square-shaped table. Rather, I meant that any round table is better (for our living room) than any square table that does not differ significantly in its other characteristics, such as height, sort of wood, finishing, price, etc. This is preference *ceteris paribus* or "everything else being equal". *Most of the preferences that we express or act upon seem to be of this type.* [Emphasis added.]

An important property of *ceteris paribus* preferential statements is their intuitive nature for users of all types. Independently of the work of philosophers in this area, reasoning about *ceteris paribus* statements has drawn the attention of AI researchers. For example, Doyle *et al.* (1991) introduced a *logic of relative desire* to treat preference statements under a *ceteris paribus* assumption. This logic bears some similarity to von Wright's (1963) logic of preferences, but supports more complicated inferences.[2] However, to the best of our knowledge, no serious attempt has been made to exploit *preferential independence* for the compact and efficient representation of such *ceteris paribus* statements. In this paper, we take steps toward structured modeling of qualitative *ceteris paribus* preferential statements.

We start by defining the notion of a (qualitative) preference relation and a number of basic preference independence concepts, followed by the introduction of CP-nets and their semantics.

### 2.1 Preference Relations

We focus our attention on single-stage decision problems with complete information, ignoring in this paper any issues that arise in multi-stage, sequential decision analysis and any considerations of risk that arise in the context of uncertainty.[3] We begin with an outline of relevant notions from decision theory. We assume that the world can be in one of a number of *states* $\mathcal{S}$ and at each state $s$ there are a number of *actions* $\mathcal{A}_s$ that can be performed. Each action, when performed at a state, has a specific *outcome* (we do not concern ourselves with uncertainty in action effects or knowledge of the state). The set of all outcomes is denoted by $\mathcal{O}$. A *preference ranking* is a total preorder $\succeq$ over the set of outcomes: $o_1 \succeq o_2$ means that outcome $o_1$ is equally or more preferred to the decision maker than $o_2$. We use $o_1 \succ o_2$ to denote the fact that outcome $o_1$ is strictly more preferred by the decision maker than $o_2$ (i.e., $o_1 \succeq o_2$ and $o_2 \not\succeq o_1$), while $o_1 \sim o_2$ denotes that the decision maker is indifferent to $o_1$ and $o_2$ (i.e., $o_1 \succeq o_2$ and $o_2 \succeq o_1$). We will use the terms preference *ordering* and *relation* interchangeably with *ranking*.

The aim of decision making under certainty is, given knowledge of a specific state, to choose the action that has the most preferred outcome. We note that the ordering $\succeq$ will vary across decision makers. For instance, two customers might have radically different preferences for computer system configurations that a sales program is helping them construct.

Often, for a state $s$, certain outcomes in $\mathcal{O}$ cannot result from any action $a \in \mathcal{A}_s$: those outcomes that can be obtained are called *feasible outcomes* (given $s$). In many instances, the mapping from states and actions to outcomes can be quite complex. In other decision scenarios, actions and outcomes may be equated: a user is allowed to directly select a feasible outcome (e.g., select a product with a desirable combination of attributes). Often states play no role (i.e., there is a single state).

One thing that makes decision problems difficult is the fact that outcomes of actions and preferences are usually not represented so directly. For example, actions may be represented as a set of constraints over a set of decision variables. We focus here on preferences. We assume a set of *variables* (or *features* or *attributes*) $\mathbf{V} = \{X_1, \ldots, X_n\}$ over which the decision maker has preferences. Each variable $X_i$ is associated with a domain $Dom(X_i) = \{x_1^i, \ldots, x_{n_i}^i\}$ of *values* it can take. An *assignment* $\mathbf{x}$ of values to a set $\mathbf{X} \subseteq \mathbf{V}$ of variables (also called an instantiation of $\mathbf{X}$) is a function that maps each variable in $\mathbf{X}$ to an element of its domain; if $\mathbf{X} = \mathbf{V}$, $\mathbf{x}$ is a *complete assignment*, otherwise $\mathbf{x}$ is called a *partial assignment*.

2. For a more detailed discussion on this issue, we refer the reader to Doyle and Wellman (1994).
3. Such issues include assigning preferences to *sequences* of outcome states, assessing uncertainty in beliefs and system dynamics, and assessing the user's attitude towards risk.

We denote the set of all assignments to $\mathbf{X} \subseteq \mathbf{V}$ by $Asst(\mathbf{X})$. If $\mathbf{x}$ and $\mathbf{y}$ are assignments to disjoint sets $\mathbf{X}$ and $\mathbf{Y}$, respectively ($\mathbf{X} \cap \mathbf{Y} = \emptyset$), we denote the combination of $\mathbf{x}$ and $\mathbf{y}$ by $\mathbf{xy}$. If $\mathbf{X} \cup \mathbf{Y} = \mathbf{V}$, we call $\mathbf{xy}$ a *completion* of assignment $\mathbf{x}$. We denote by $Comp(\mathbf{x})$ the set of completions of $\mathbf{x}$. Complete assignments correspond directly to the outcomes over which a user possesses preferences. For any outcome $o$, we denote by $o[X]$ the value $x \in Dom(X)$ assigned to variable $X$ by that outcome.

Given a problem defined over $n$ variables with domains $Dom(X_1), \ldots, Dom(X_n)$, there are $|Dom(X_1)| \times \cdots \times |Dom(X_n)|$ assignments. Thus direct assessment of a preference function is usually infeasible due to the exponential number of outcomes. Fortunately, a preference function can be specified (or partially specified) concisely if it exhibits sufficient structure. We describe certain standard types of structure here, referring to Keeney and Raiffa (1976) for a detailed description of these (and other) structural forms and a discussion of their implications. A set of variables $\mathbf{X}$ is *preferentially independent* of its complement $\mathbf{Y} = \mathbf{V} - \mathbf{X}$ iff, for all $\mathbf{x}_1, \mathbf{x}_2 \in Asst(\mathbf{X})$ and $\mathbf{y}_1, \mathbf{y}_2 \in Asst(\mathbf{Y})$, we have

$$\mathbf{x}_1\mathbf{y}_1 \succeq \mathbf{x}_2\mathbf{y}_1 \text{ iff } \mathbf{x}_1\mathbf{y}_2 \succeq \mathbf{x}_2\mathbf{y}_2.$$

In other words, the structure of the preference relation over assignments to $\mathbf{X}$, when all other variables are held fixed, is the same no matter what values these other variables take. If the relation above holds, we say $\mathbf{x}_1$ is preferred to $\mathbf{x}_2$ *ceteris paribus*. Thus, one can assess the relative preferences over assignments to $\mathbf{X}$ once, knowing these preferences do not change as other attributes vary.

We define conditional preferential independence analogously. Let $\mathbf{X}$, $\mathbf{Y}$, and $\mathbf{Z}$ be nonempty sets that partition $\mathbf{V}$. $\mathbf{X}$ is *conditionally preferentially independent* of $\mathbf{Y}$ given an assignment $\mathbf{z}$ to $\mathbf{Z}$ iff, for all $\mathbf{x}_1, \mathbf{x}_2 \in Asst(\mathbf{X})$ and $\mathbf{y}_1, \mathbf{y}_2 \in Asst(\mathbf{Y})$, we have

$$\mathbf{x}_1\mathbf{y}_1\mathbf{z} \succeq \mathbf{x}_2\mathbf{y}_1\mathbf{z} \text{ iff } \mathbf{x}_1\mathbf{y}_2\mathbf{z} \succeq \mathbf{x}_2\mathbf{y}_2\mathbf{z}.$$

In other words, $\mathbf{X}$ is preferentially independent of $\mathbf{Y}$ when $\mathbf{Z}$ is assigned $\mathbf{z}$. If $\mathbf{X}$ is conditionally preferentially independent of $\mathbf{Y}$ for all $\mathbf{z} \in Asst(\mathbf{Z})$, then $\mathbf{X}$ is *conditionally preferentially independent* of $\mathbf{Y}$ given the set of variables $\mathbf{Z}$.

Note that the *ceteris paribus* component of these definitions ensures that the statements one makes are relatively weak. In particular, they do not imply a stance on specific value tradeoffs. Consider two variables $A$ and $B$ that are preferentially independent, so that the preferences for values of $A$ and $B$ can be assessed separately; for instance, suppose $a_1 \succ a_2$ and $b_1 \succ b_2$. Clearly, $a_1b_1$ is the most preferred outcome and $a_2b_2$ is the least; but if feasibility constraints make $a_1b_1$ impossible, we must be satisfied with one of $a_1b_2$ or $a_2b_1$. We cannot tell which is most preferred using these separate assessments. However, under stronger conditions (e.g., *additive preferential independence*) one can construct an additive value function in which weights are assigned to different attributes (or attribute groups). Such a decomposition of a preference function allows one to identify the most preferred outcomes rather readily, and this, as well as some other special forms of preference structure, are especially appropriate when attributes take on numerical values. For an extensive discussion of various special forms of preference functions we refer to Keeney and Raiffa (1976), as well as Bacchus and Grove (1995, 1996) and Shoham (1997).

### 2.2 CP-Networks

Our representation for preferences is graphical in nature, and exploits conditional preferential independence in structuring preferences of a user. The model is similar to a Bayesian network (Pearl, 1988) on the surface; however, the nature of the relation between nodes within a network is generally quite weak (e.g., compared with the probabilistic relations in Bayes nets). Others have defined graphical representations of preference relations; for instance Bacchus and Grove (1995, 1996) have shown some strong results pertaining to undirected graphical representations of additive independence. Our representation and semantics is rather distinct, and our main aim in using the graph is to capture statements of *qualitative* conditional preferential independence. We note that reasoning about *ceteris paribus* statements has been explored in AI (Doyle et al., 1991; Wellman & Doyle, 1991; Doyle & Wellman, 1994), though not in the context of network representations or exploiting preferential independence in a computational fashion.

For each variable $X_i$, we ask the user to identify a set of *parent* variables $Pa(X_i)$ that can affect her preference over various values of $X_i$. That is, given a particular value assignment to $Pa(X_i)$, the user should be able to determine a preference order for the values of $X_i$, all other things being equal. Formally, given $Pa(X_i)$ we have that $X_i$ is conditionally preferentially independent of $\mathbf{V} - (Pa(X_i) \cup \{X_i\})$. Given this information, we ask the user to explicitly specify her preferences over the values of $X_i$ for all instantiations of the variable set $Pa(X_i)$. We use the above information to create an annotated directed graph in which nodes stand for the problem variables, and every node $X_i$ has $Pa(X_i)$ as its immediate ancestors. The node $X_i$ is annotated with a *conditional preference table* (CPT) describing the user's preferences over the values of the variable $X_i$ given every combination of parent values. In other words, letting $Pa(X_i) = \mathbf{U}$, for each assignment $\mathbf{u} \in Asst(\mathbf{U})$, we assume that a total preorder $\succeq^i_{\mathbf{u}}$ is provided over the domain of $X_i$: for any two values $x$ and $x$, either $x \succ^j_{\mathbf{u}} x'$, $x' \succ^j_{\mathbf{u}} x$, or $x \sim^j_{\mathbf{u}} x'$. For simplicity of presentation, we ignore indifference in our algorithms. Though treatment of indifference is straightforward semantically, consistency of arbitrary networks with indifference cannot be assumed, as we discuss in Section 2.5. Likewise, we assume that CPTs for all variables are fully specified, though we discuss partially specified CPTs in Section 6.

We call these structures *conditional preference networks* or *CP-networks* (CP-nets, for short).

**Definition 1** A *CP-net* over variables $\mathbf{V} = \{X_1, \ldots, X_n\}$ is a directed graph $G$ over $X_1, \ldots, X_n$ whose nodes are annotated with conditional preference tables $CPT(X_i)$ for each $X_i \in \mathbf{V}$. Each conditional preference table $CPT(X_i)$ associates a total order $\succ^i_{\mathbf{u}}$ with each instantiation $\mathbf{u}$ of $X_i$'s parents $Pa(X_i) = \mathbf{U}$.

We illustrate the CP-net semantics and some of its consequences with several small examples. For ease of presentation, all variables in these examples are boolean, though our semantics is defined for features with arbitrary finite domains.

**Example 1** (My Dinner I) Consider the simple CP-net in Figure 1(a) that expresses my preference over dinner configurations. This network consist of two variables $S$ and $W$, standing for the soup and wine, respectively. Now, I strictly prefer fish soup ($S_f$) to vegetable soup ($S_v$), while my preference between red ($W_r$) and white ($W_w$) wine is conditioned on the soup to be served: I prefer red wine if served a vegetable soup, and white wine if served a fish soup.

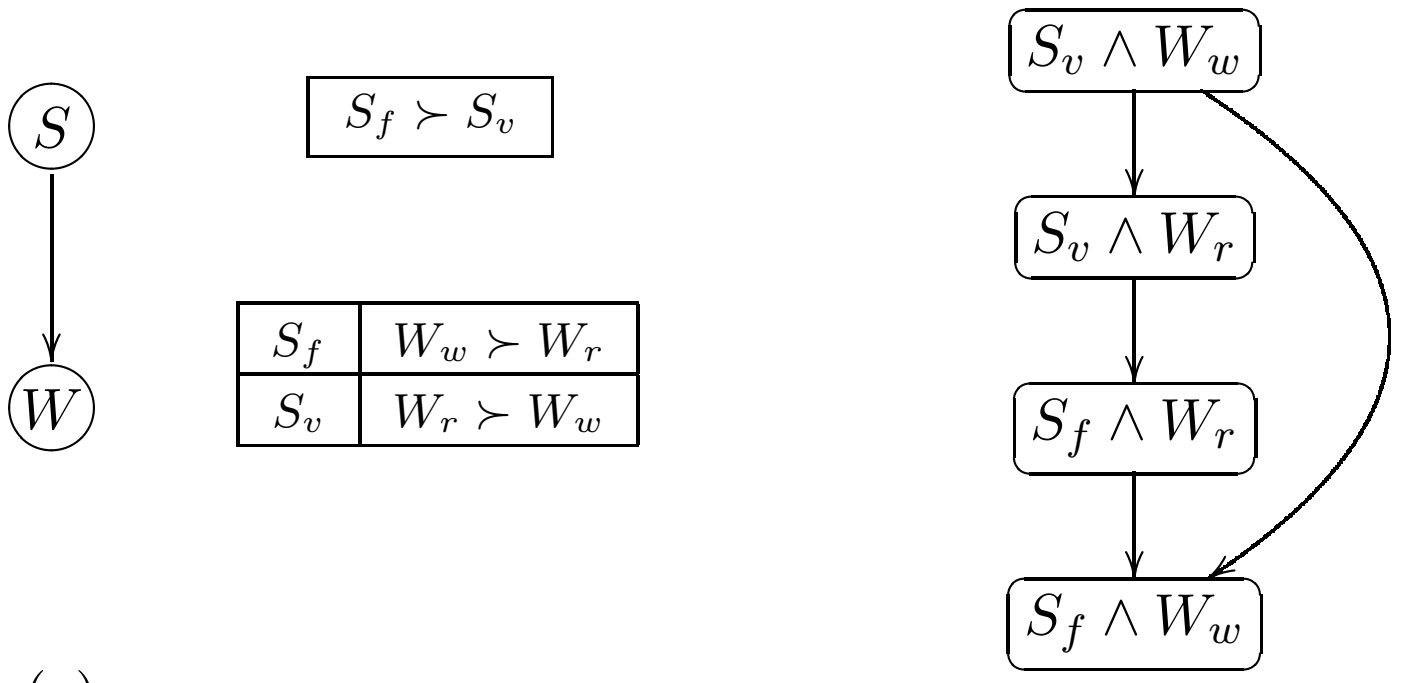

(a) (b)

Figure 1: (a) CP-net for "My Dinner I": Soup and Wine; (b) the induced preference graph.

Figure 1(b) shows the *preference graph* over outcomes induced by this CP-net. An arc in this graph directed from outcome $o_i$ to $o_j$ indicates that a preference for $o_j$ over $o_i$ can be determined directly from one of the CPTs in the CP-net. For example, the fact that $S_v \wedge W_r$ is preferred to $S_v \wedge W_w$ (as indicated by the direct arc between them) is a direct consequence of the semantics of $CPT(W)$. The top element ($S_v \wedge W_w$) is the worst outcome while the bottom element ($S_f \wedge W_w$) is the best. □

**Example 2** (My Dinner II) Figure 2(a) extends the *chain* CP-net of Example 1 by adding the main course $M$ as another variable. In this example, my preference over the options for the main course is clear: I strictly prefer a meat course $M_{mc}$ to a fish course $M_{fc}$. In addition, I prefer not to have two fish courses in one dinner; thus my preference between vegetable and fish soup is conditioned on the main course: I prefer to open with fish soup if the main course is meat, and with vegetable soup if the main course is fish. As in Example 1, Figure 2(b) shows the corresponding induced preference graph over outcomes. □

**Example 3** (Evening Dress) Figure 3(a) illustrates another CP-net that expresses my preferences for evening dress. It consists of three variables $J$, $P$, and $S$, standing for the jacket, pants, and shirt, respectively. I unconditionally prefer black to white as a color for both the jacket and the pants, while my preference between the red and white shirts is conditioned on the *combination* of jacket and pants: if they have the same color, then a white shirt will make my outfit too colorless, thus I prefer a red shirt. Otherwise, if the jacket and the pants are of different colors, then a red shirt will probably make my outfit too flashy, thus I prefer a white shirt. Figure 3(b) shows the corresponding preference graph. □

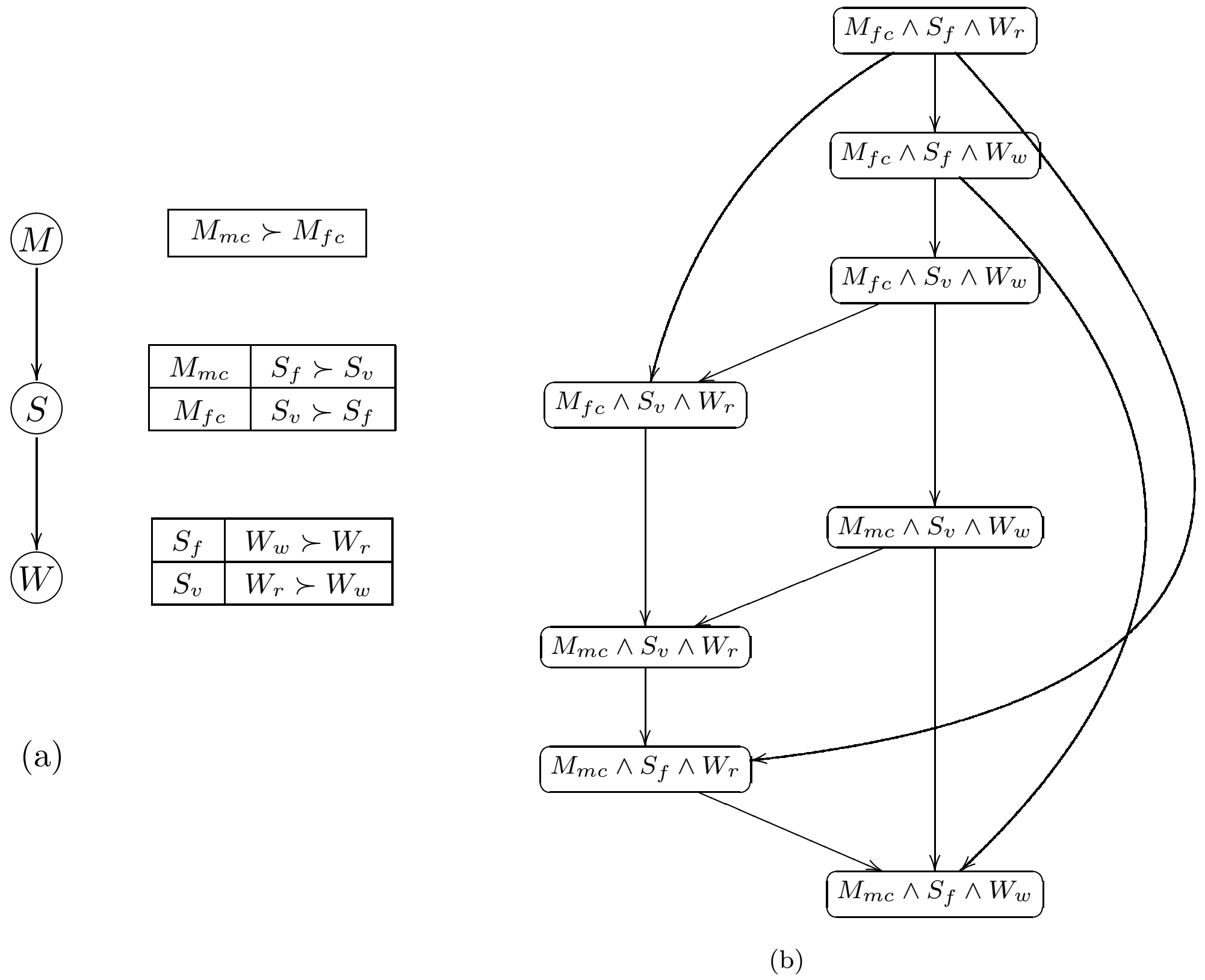

Figure 2: (a) CP-net for "My Dinner II"; (b) the induced preference graph.

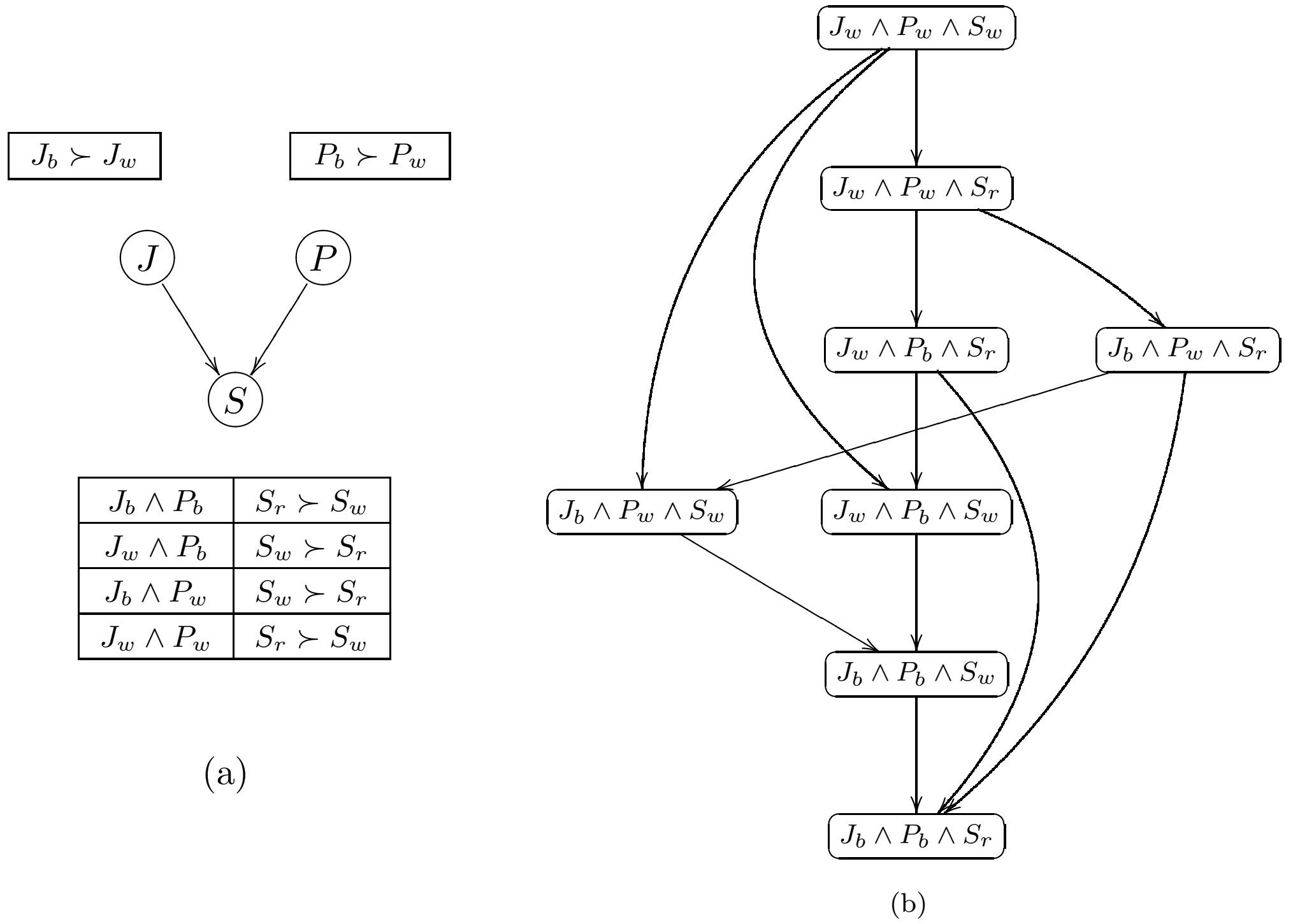

Figure 3: (a) CP-Net for "Evening Dress": Jacket, Pants and Shirt; (b) the induced preference graph.

### 2.3 Semantics

The semantics of a CP-net is straightforward. It is defined in terms of the set of preference rankings that are consistent with the set of preference constraints imposed by its CPTs.

**Definition 2** Let $N$ be a CP-net over variables $\mathbf{V}$, $X_i \in \mathbf{V}$ be some variable, and $\mathbf{U} \subset \mathbf{V}$ be the parents of $X_i$ in $N$. Let $\mathbf{Y} = \mathbf{V} - (\mathbf{U} \cup \{X_i\})$. Let $\succ^i_{\mathbf{u}}$ be the ordering over $Dom(X_i)$ dictated by $CPT(X_i)$ for any instantiation $\mathbf{u} \in Asst(\mathbf{U})$ of $X_i$'s parents. Finally let $\succ$ be a preference ranking over $Asst(\mathbf{V})$.

A preference ranking $\succ$ *satisfies* $\succ^i_{\mathbf{u}}$ iff we have—for all $\mathbf{y} \in Asst(\mathbf{Y})$ and all $x, x' \in Dom(X_i)$— $\mathbf{y}x\mathbf{u} \succ \mathbf{y}x'\mathbf{u}$ whenever $x \succ^i_{\mathbf{u}} x'$. A preference ranking $\succ$ *satisfies* the CPT $CPT(X_i)$ iff it satisfies $\succ^i_{\mathbf{u}}$ for each $\mathbf{u} \in Asst(\mathbf{U})$. A preference ranking $\succ$ *satisfies* the CP-net $N$ iff is satisfies $CPT(X_i)$ for each variable $X_i$.

A CP-net $N$ is *satisfiable* iff there is some preference ranking $\succ$ that satisfies it.

Thus a network $N$ is satisfied by $\succ$ iff $\succ$ satisfies each of the conditional preferences expressed in the CPTs of $N$ under the *ceteris paribus* interpretation.

**Theorem 1** *Every acyclic CP-net is satisfiable.*

**Proof:** We prove this constructively by building a satisfying preference ordering. This proof is by induction on the number of variables. The theorem trivially holds for one variable, as the total ordering is specified directly by the CP-net.

Suppose the theorem holds for all CP-nets with fewer than $n$ variables. Let $N$ be a network with $n$ variables. If $N$ is acyclic, there is at least one variable with no parents; let $X$ be such a variable. Let $x_1 \succ x_2 \succ \ldots \succ x_k$ be the ordering over $Dom(X)$ dictated by $CPT(X)$. For each $x_i$, construct a CP-net, $N_i$, with the $n-1$ variables $\mathbf{V} - \{X\}$ by removing $X$ from the initial CP-net, and for each variable $Y$ that is a child of $X$, revising its CPT by restricting each row to $X = x_i$. By the inductive hypothesis, we can construct a preference ordering $\succ_i$ for each of the reduced CP-nets $N_i$.

We can now construct a preference ordering for the original network $N$ as follows. We rank every outcome with $X = x_i$ as preferred to any outcome with $X = x_j$ if $x_i \succ x_j$ in $CPT(X)$. For any outcomes with identical values $x_i$ of $X$, we rank them according to the ordering $\succ_i$ associated with $N_i$ (ignoring the value of $X$). It is easy to see that this preference ordering satisfies $N$. $\square$

For example, consider the CP-net of Example 1 (Figure 1). Somewhat surprisingly, the information captured by this network is sufficient to totally order the outcomes:

$$S_f \wedge W_w \;\succ\; S_f \wedge W_r \;\succ\; S_v \wedge W_r \;\succ\; S_v \wedge W_w$$

since this is the only ranking that satisfies this CP-net. However, this need not be the case—in general, a satisfiable CP-net can be satisfied by more than one ranking. For instance, consider the CP-net in Figure 4.[4] There are two rankings that satisfy this network:

$$abc \succ ab\bar{c} \succ a\bar{b}\bar{c} \succ a\bar{b}c \succ \bar{a}\bar{b}\bar{c} \succ \bar{a}\bar{b}c \succ \bar{a}bc \succ \bar{a}b\bar{c}$$
$$abc \succ ab\bar{c} \succ a\bar{b}\bar{c} \succ \bar{a}\bar{b}\bar{c} \succ a\bar{b}c \succ \bar{a}\bar{b}c \succ \bar{a}bc \succ \bar{a}b\bar{c}$$

Preferential entailment in a CP-net is defined in a standard way.

**Definition 3** Let $N$ be a CP-net over variables $\mathbf{V}$, and $o, o' \in Asst(\mathbf{V})$ be any two outcomes. *$N$ entails $o \succ o'$* (i.e., that outcome $o$ is preferred to $o'$), written $N \models o \succ o'$, iff $o \succ o'$ holds in every preference ordering that satisfies $N$.

**Lemma 2** *Preferential entailment is transitive. That is, if $N \models o \succ o'$ and $N \models o' \succ o''$ then $N \models o \succ o''$.*

**Proof:** If $N \models o \succ o'$ and $N \models o' \succ o''$ then $o \succ o'$ and $o' \succ o''$ in all preference rankings satisfying $N$. As each of these rankings is transitive, we must have $o \succ o''$ in all satisfying rankings. $\square$

For example, consider the CP-net $N$ in Figure 4(a) and the following three outcomes: $o = ab\bar{c}$, $o' = a\bar{b}\bar{c}$, and $o'' = a\bar{b}c$. The outcomes $o$ and $o'$ assign the same values to all variables except of $B$. In addition, given the value of $Pa(B) = \{A\}$ in $o$ and $o'$, the value of $B$ in $o$ ($B = b$) is preferred to the value of $B$ ($B = \bar{b}$) in $o'$, all else being equal. Therefore, we have that $N \models o \succ o'$. In the case of $o'$ and $o''$, the same argument with respect to the

4. This is the network from Example 2 ("My Dinner II") with the variables renamed.

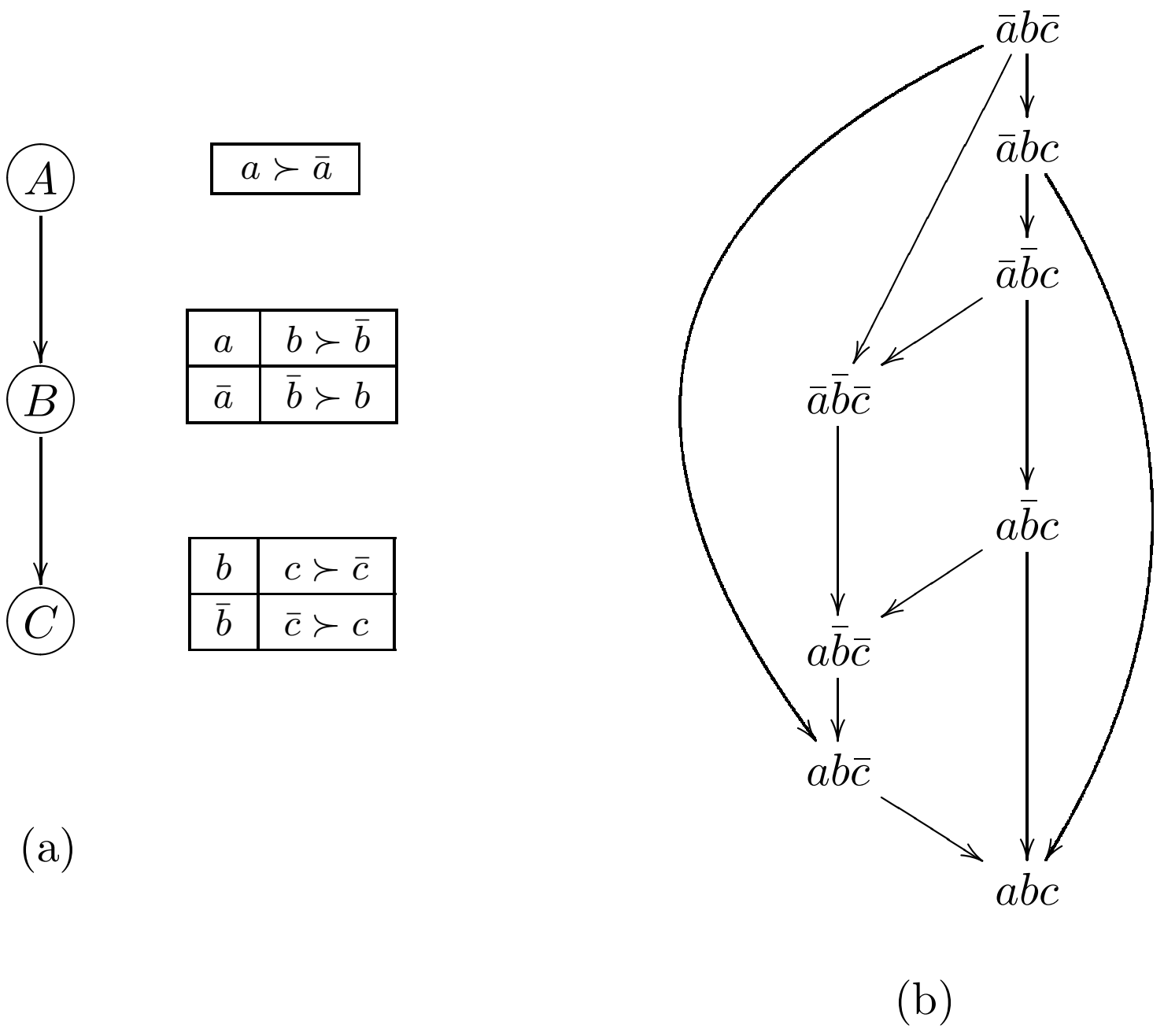

Figure 4: A simple chain-structured CP-network.

variable $C$ will show that $N \models o' \succ o''$ as well. Observe that $o \succ o''$ cannot be derived directly from the CPTs of $N$. However, from Lemma 2, it follows that this relation can be inferred by taking the transitive closure of the direct relations $o \succ o'$ and $o' \succ o''$.

Notice that, given a CP-net, we can assess each outcome in terms of the conditional preferences it violates. For example, in the CP-net of Example 1: the outcome $S_f \wedge W_w$ violates none of the preference constraints; $S_f \wedge W_r$ violates the conditional preference for $W$; $S_v \wedge W_r$ violates the preference for $S$; and $S_v \wedge W_w$ violates both. Somewhat surprisingly, the *ceteris paribus* semantics implicitly ensures that violating the preference for $S$ is worse than violating that for $W$, since $S_f \wedge W_r \succ S_v \wedge W_r$. That is, the parent preferences have higher priority than the child preferences. This property has important implications for inference as we will see below.

## 2.4 Cyclic Networks

As mentioned, nothing in the semantics of the CP-net model forces it to be acyclic. However, according to Theorem 1, the acyclicity of the network automatically confers an important property to the model: the network is satisfiable (i.e., there exists a preference ordering that satisfies all *ceteris paribus* preference assertions imposed by the CPTs).

For cyclic CP-nets, the situation is much more complicated. For example, consider a binary-valued cyclic CP-net structure in Figure 5(e). If the CPTs for this network are specified as in Figure 5(a), then the induced preference graph (see Figure 5(b)) can be extended to a complete preference ordering consistently. However, if the CPTs are specified as in Figure 5(c), then the network is unsatisfiable (the induced preference graph, shown in Figure 5(d) cannot be completed consistently). This example shows that the consistency of cyclic CP-nets is not guaranteed, and depends on the actual nature of the CPTs.

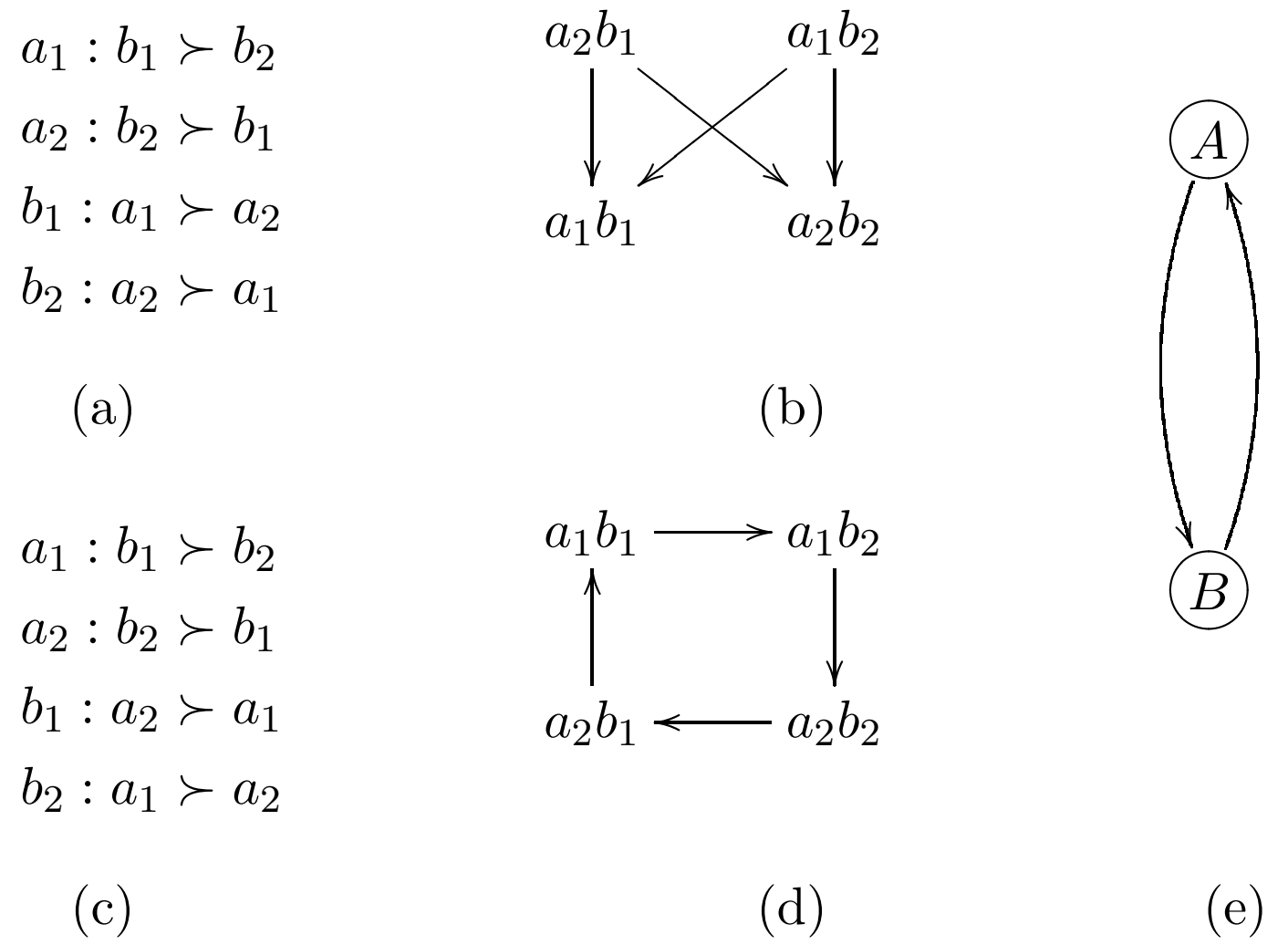

Figure 5: Examples of a satisfiable and an unsatisfiable cyclic CP-net over binary variables.

Recently, initial results on consistency testing for cyclic CP-nets were presented by Domshlak and Brafman (2002a). In particular, a wide class of cyclic, binary-valued CP-nets was identified to be efficiently testable for consistency. However, these results cover only part of the spectrum, and further research on cyclic CP-nets is needed.

Beyond the open computational questions that cyclic CP-nets raise, their usefulness requires further analysis. One can argue that it is possible to cluster the variables to preserve acyclicity. Although this approach is technically feasible and probably useful in many domains, it cannot provide a general solution. First, such clustering will affect the space requirements of problem description and thus, it will generally degrade the efficiency of reasoning about preferences. Second, in certain domains, it may be more natural to express cyclic preferences even if an acyclic representation could be used. For example, this seems to be the case in work on preference-based presentation of web page content (Domshlak, Brafman, & Shimony, 2001), where it is argued that the preferred presentation of a certain component of a web page may depend on the presentation of its neighbors in the page, whose preferred presentation depend on its presentation, and so on.

One could argue that preferences naturally exhibit cyclic structure and that acyclic nets are of theoretical interest only. Our experience indicates the opposite. Acyclic CP-nets are shown to be effective and natural in the above-mentioned work on web page presentation (Domshlak et al., 2001), as well as in a related project that deals with the presentation of multi-media content in a medical domain (Gudes, Domshlak, & Orlov, 2002)—a more extensive example from this latter domain is presented in the next section. Moreover, in other domains, we have found it difficult to generate intuitively reasonable cyclic networks. This is due to the fact that a cycle implies that all variables in it are equally important.

Typically, this is not the case. Thus, because of the apparent utility of acyclic networks, the fact that we can use composite variables made by clustering primitive variables, and the additional complexity involved in cyclic networks, we consider only acyclic CP-nets in the remainder of this paper. However, further investigation of cyclic CP-nets, as well as a characterization of the different classes of utility functions that can be represented by cyclic and acyclic networks, remains of interest.

### 2.5 Indifference

We have so far assumed that the preference constraints in each CPT of a CP-net totally order the outcomes of the variable in question. Specifically, for any variable $X_i$ with parents $\mathbf{U}$, and any $\mathbf{u} \in Asst(\mathbf{U})$, we assume that $\succ^i_{\mathbf{u}}$ is a total order over $Dom(X_i)$. The general definition of a CP-net can allow an arbitrary total preorder $\succeq^i_{\mathbf{u}}$ over $Dom(X_i)$, thus allowing the user to express indifference between two values of variable $X$, say $x$ and $x'$, given $\mathbf{u}$. We denote this by $x \sim x'$ in $CPT(X)$.

It turns out that the *ceteris paribus* semantics is quite strong when we say that two variable values are equally preferred.

**Example 4** Consider the following two CPTs for a network over variables $A$ and $B$, with $A$ being a parent of $B$:

$$a \sim \overline{a};$$
$$a : b \succ \overline{b}; \quad \overline{a} : \overline{b} \succ b;$$

These assert that the user is indifferent between $a$ and $\overline{a}$, but should $a$ hold, prefers $b$, and should $\overline{a}$ hold, prefers $\overline{b}$. We can derive the following preferences over outcomes:

$$ab \succ a\overline{b} \succeq \overline{a}\overline{b} \succ \overline{a}b \succeq ab$$

These statements are not consistent with any preference ranking, hence this network is not satisfiable. One way to interpret this is that if someone really did have the preferences:

$$a : b \succ \overline{b}; \quad \overline{a} : \overline{b} \succ b;$$

they cannot be indifferent between $a$ and $\overline{a}$, *ceteris paribus*. □

This points to a potential difficulty with the use of indifference in CP-nets. One must be careful not to express indifference between two values of a variable ($A$ in this case), yet express a (strict) conditional preference for a child of that variable ($B$) that depends on the values for which the user is indifferent. Intuitively, in this case, it seems that the user is expressing the fact that they would like the value of $B$ to match that of $A$ (with respect to their "sign"), but intends no preference for $ab$ over $\overline{a}\overline{b}$ (or vice versa). If this is the case, then making $A$ a parent of $B$ expresses that the preference for $B$ is subsidiary to that of $A$, which is not the intent. In this case, either a cyclic network (indeed the satisfiable network discussed in Section 2.4) or the clustering of variables $A$ and $B$ seems appropriate.

Despite this, indifference can be used safely as follows. Let $X_i$ be any variable in network $N$ with parents $\mathbf{U}$, and let $X_j$ be any child of $X_i$. Let $\mathbf{Y}$ denote the remaining parents of $X_j$ (those excluding $X_i$). Suppose that for some $\mathbf{u} \in Asst(\mathbf{U})$, and $x, x' \in Dom(X_i)$, we have $x \sim x'$ in $\succeq^i_{\mathbf{u}}$. Then as long as the local orderings in $CPT(X_j)$ for a fixed instantiation of $\mathbf{Y}$ are identical whether $x$ or $x'$ holds, then the network $N$ is satisfiable. More precisely, if $\succeq^j_{x\mathbf{y}} = \succeq^j_{x'\mathbf{y}}$ for each $\mathbf{y} \in Asst(\mathbf{Y})$, then network $N$ is satisfiable. Thus, if we are indifferent between $x$ and $x'$, then our preferences over values of $X_i$'s children, should exhibit indifference whether the context includes $x$ or $x'$.[5]

For simplicity of presentation, for the remainder of the paper we continue to assume that preference constraints in each CPT of a CP-net totally order the outcomes of the variable in question. However, in Section 6, we do discuss the applicability of our results to satisfiable CP-nets that capture statements of preferential indifference.

## 3. Outcome Optimization

One of the principal properties of CP-nets is that, given a CP-net $N$, we can easily determine the best outcome among those preference rankings that satisfy $N$. We call such a query an *outcome optimization* query. This turns to be a simple task in CP-nets.

### 3.1 An Algorithm for Outcome Optimization

Intuitively, to generate an optimal outcome we simply need to sweep through the network from top to bottom (i.e., from ancestors to descendents) setting each variable to its most preferred value given the instantiation of its parents. Indeed, while the network does not generally determine a unique ranking, it does determine a unique best outcome (assuming no indifference). More generally, suppose we are given evidence constraining possible outcomes in the form of an instantiation $\mathbf{z}$ of some subset $\mathbf{Z} \subseteq \mathbf{V}$ of the network variables. Determining the best completion of $\mathbf{z}$ (that is, the best outcome consistent with $\mathbf{z}$) can be achieved in a similar fashion, as we now outline.

Outcome optimization queries can be answered using the following *forward sweep* procedure, taking time linear in the number of variables. Assume a partial instantiation $\mathbf{z} \in Asst(\mathbf{Z})$, and the goal of determining the (unique) $o \in Comp(\mathbf{z})$ such that $N \models o \succ o'$ for all $o' \in Comp(\mathbf{z}) - \{o\}$. This can be effected by a straightforward sweep through the network. Let $X_1, \ldots, X_n$ be any topological ordering of the network variables. We set $\mathbf{Z} = \mathbf{z}$, and instantiate each $X_i \notin \mathbf{Z}$ in turn to its maximal value given the instantiation of its parents. This procedure exploits the considerable power of both the *ceteris paribus* semantics and the graphical modeling of the preferential statements to easily find an optimal outcome given certain observed evidence (or imposed conjunctive constraints).

**Lemma 3** *The forward sweep procedure constructs the most preferred outcome in* $Comp(\mathbf{z})$.

**Proof:** Let $\mathbf{v_z}$ be any outcome in the set of completions of $\mathbf{z}$. Define a sequence of outcomes $\mathbf{v}_i$, $0 \le i \le n$, as follows: (a) $\mathbf{v}_0 = \mathbf{v_z}$; (b) if $X_i \notin \mathbf{Z}$, $\mathbf{v}_i$ is constructed by setting the value of $X_i$ to its most preferred value given the instantiation of its parents in $\mathbf{v}_{i-1}$, with all other variables retaining their values from $\mathbf{v}_{i-1}$; (c) if $X_i \in \mathbf{Z}$, then $\mathbf{v}_i = \mathbf{v}_{i-1}$. By construction, $\mathbf{v}_i \succeq \mathbf{v}_{i-1}$. The last outcome $\mathbf{v}_n$ is precisely that constructed by the forward

5. This restriction can be relaxed somewhat if we take into account the fact that some of $X_j$'s parents could lie in the set $\mathbf{U}$, in which case these rankings need not agree for *every* indifference pair $x$ and $x'$.

sweep algorithm. Notice that we arrive at the same outcome irrespective of our starting point $\mathbf{v_z}$ (by assumption, there can be no ties). Since $\mathbf{v}_n \succeq \mathbf{v_z}$ for any outcome $\mathbf{v_z}$ consistent with the evidence, the forward sweep procedure yields the optimal outcome. □

### 3.2 An Example Application

We now turn to an illustration of the use of CP-nets in the context of a CP-net based system for adaptive multimedia document presentation. Applications based on this system for the presentation of web-based content and multi-media medical data were recently presented by Domshlak et al. (2001) and Gudes et al. (2002). Through this example we demonstrate the simplicity of preference specification using CP-nets, the utility of acyclic networks, and the use of the optimization algorithm described above.

The system consists of two tools—the authoring tool, and the viewing tool. The central part of the authoring tool is a module for the specification of a CP-net corresponding to the created and/or edited multimedia document. Using this CP-net, a content provider express her preferences regarding the presentation of the document content. For example, the content provider may prefer that some material be presented if and only if some other material is not presented. The result of such preference-based multimedia document design is a meta-document specifying both *what to present* and *how to present it*.

The description of the content provider's preferences, as captured by an acyclic CP-net, becomes a static part of the document, and sets the parameters for its initial presentation. Given such a document, the viewing tool is responsible for reasoning about these preferences; specifically, it must determine an optimal reconfiguration of the document context after interaction of the viewer with the document. In this process, the user's $k$ most recent content choices are viewed as constraints of the form *"these items must appear as I specified"*. Subject to these constraints, an optimal document presentation with respect to the content provider's CP-net must then be generated. Thus, for each particular session, the actual presentation changes dynamically based on the user's choices. More precisely, whenever new user input is obtained, the optimization algorithm constructs the best presentation of all document components with respect to the content provider's preferences among those presentations that conform to the user's recent viewing choices. This process uses the forward sweep procedure described above.

**Example 5** (Multimedia Document) Consider a medical record that consists of six components: two components correspond to a set of medical tests conducted in 2001—an X-ray image and textual notes of a physician—and four components correspond to a set of medical tests from 2002—a CT (computerized tomography) image, an X-ray image, a graph illustrating results of electromyography, and textual notes of a physician. For the purposes of illustration, we assume the preferences of a content provider (e.g., the latter physician) over the presentation options of these components can be captured using the CP-net shown in Figure 7. The specific details of the preferences—the nature of the preferential dependencies and the precise details of the CPTs are summarized as follows:

- CT-image [CT image, 2002] consist of four CT images of different parts of the body, and it is shown in Figure 6(a). There are six presentation options for CT-image: it can be either completely presented ($\mathsf{ct}_{plain}$), or completely hidden ($\mathsf{ct}_{hide}$), or presented by a zoom-in on one of its four parts ($\mathsf{ct}_{lt}$, $\mathsf{ct}_{rt}$, $\mathsf{ct}_{lb}$, and $\mathsf{ct}_{rb}$, standing for left-top, right-top, left-bottom, and right-bottom parts, respectively). The physician's preference over the presentation options of CT-image is unconditional:

$$\mathsf{ct}_{hide} \succ \mathsf{ct}_{lt} \succ \mathsf{ct}_{rt} \succ \mathsf{ct}_{lb} \succ \mathsf{ct}_{rb} \succ \mathsf{ct}_{plain}$$

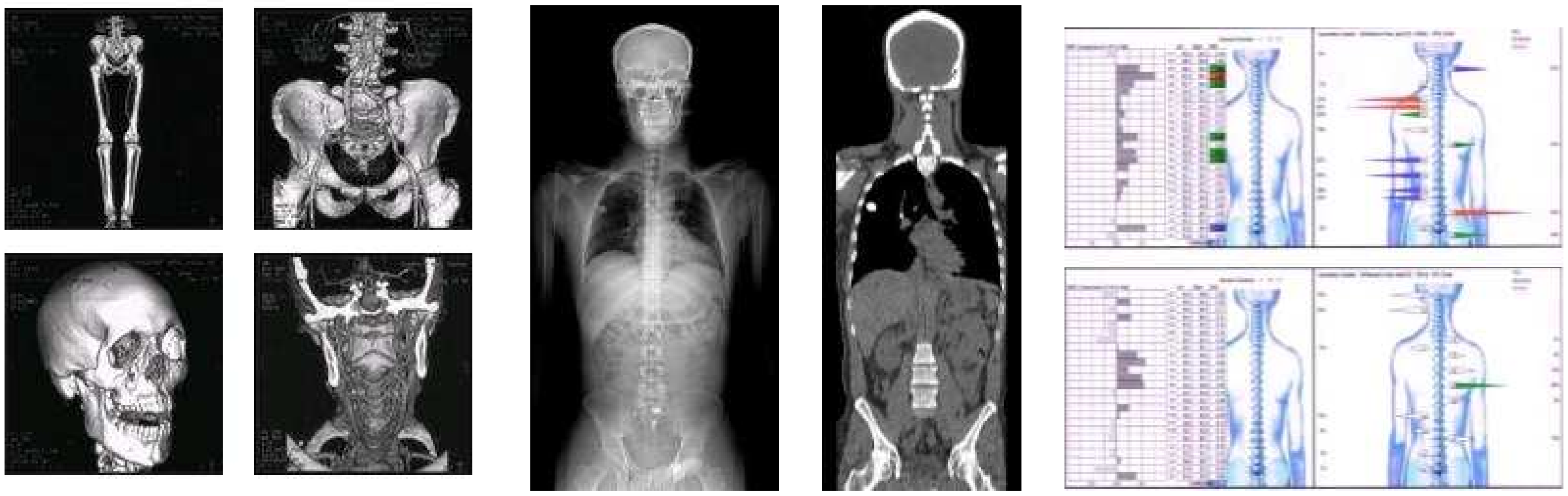

(a) (b) (c) (d)

Figure 6: Document components in Multimedia Document example.

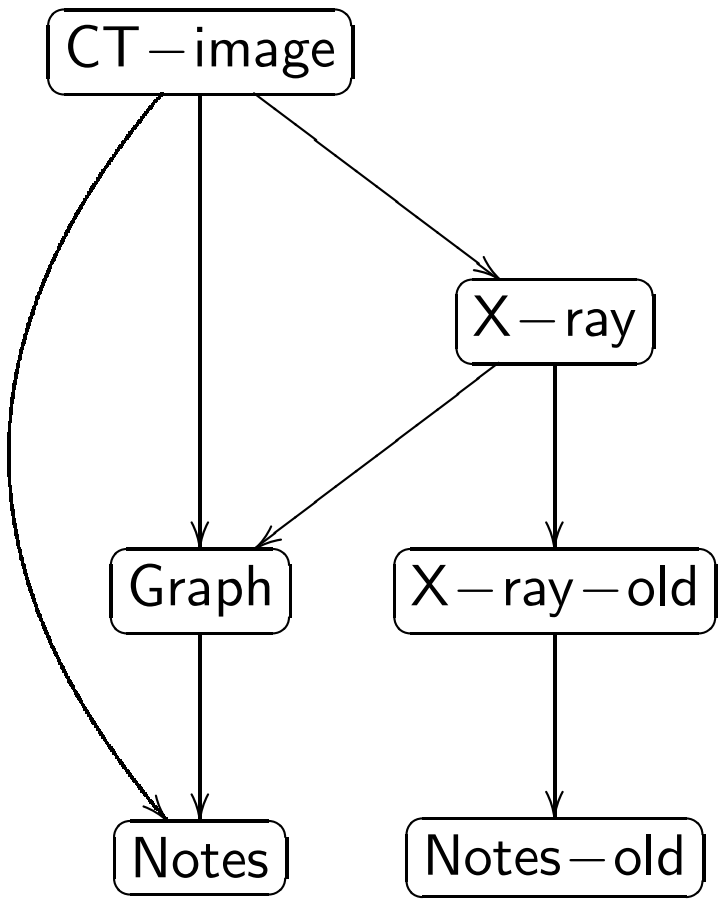

Figure 7: Multimedia Document CP-net.

- X-ray [X-ray, 2002] can be either hidden ($\mathsf{xray}_{hide}$), or presented as is ($\mathsf{xray}_{plain}$), or presented after a segmentation ($\mathsf{xray}_{segm}$); the image and its segmentation are depicted in Figures 6(b) and 6(c), respectively. The preference over the presentation options of X-ray depends on the presentation of CT-image:

| | |
|---|---|
| $\mathsf{ct}_{plain}$ | $\mathsf{xray}_{hide} \succ \mathsf{xray}_{plain} \succ \mathsf{xray}_{segm}$ |
| $\neg(\mathsf{ct}_{plain} \vee \mathsf{ct}_{hide})$ | $\mathsf{xray}_{plain} \succ \mathsf{xray}_{segm} \succ \mathsf{xray}_{hide}$ |
| $\mathsf{ct}_{hide}$ | $\mathsf{xray}_{segm} \succ \mathsf{xray}_{plain} \succ \mathsf{xray}_{hide}$ |

- $\mathsf{Graph}$ [Electromyography, 2002] is shown in Figure 6(d), and it can be either presented ($\mathsf{graph}_{plain}$), or hidden ($\mathsf{graph}_{hide}$). The preference over the presentation options of $\mathsf{Graph}$ depends on the presentation of both $\mathsf{CT\text{-}image}$ and $\mathsf{X\text{-}ray}$:

| | |
|---|---|
| $(\mathsf{ct}_{lt} \vee \mathsf{ct}_{rt} \vee \mathsf{ct}_{lb} \vee \mathsf{ct}_{rb}) \vee \mathsf{xray}_{segm}$ | $\mathsf{graph}_{plain} \succ \mathsf{graph}_{hide}$ |
| otherwise | $\mathsf{graph}_{hide} \succ \mathsf{graph}_{plain}$ |

- $\mathsf{Notes}$ [Textual notes, 2002] can be either fully presented ($\mathsf{notes}_{plain}$), or summarized ($\mathsf{notes}_{summ}$), or omitted all together ($\mathsf{notes}_{hide}$). The preference over the presentation options of $\mathsf{Notes}$ depends on the presentation of both $\mathsf{CT\text{-}image}$ and $\mathsf{Graph}$:

| | |
|---|---|
| $\mathsf{ct}_{hide}$ | $\mathsf{notes}_{hide} \succ \mathsf{notes}_{summ} \succ \mathsf{notes}_{plain}$ |
| $\neg(\mathsf{ct}_{hide}) \wedge \mathsf{graph}_{plain}$ | $\mathsf{notes}_{summ} \succ \mathsf{notes}_{plain} \succ \mathsf{notes}_{hide}$ |
| otherwise | $\mathsf{notes}_{hide} \succ \mathsf{notes}_{summ} \succ \mathsf{graph}_{plain}$ |

- $\mathsf{X\text{-}ray\text{-}old}$ [X-ray, 2001] can be either hidden ($\mathsf{xray\!-\!old}_{hide}$), or presented as is ($\mathsf{xray\!-\!old}_{plain}$); the image is depicted below. The preference over the presentation options of $\mathsf{X\text{-}ray\text{-}old}$ depends on the presentation of $\mathsf{X\text{-}ray}$:

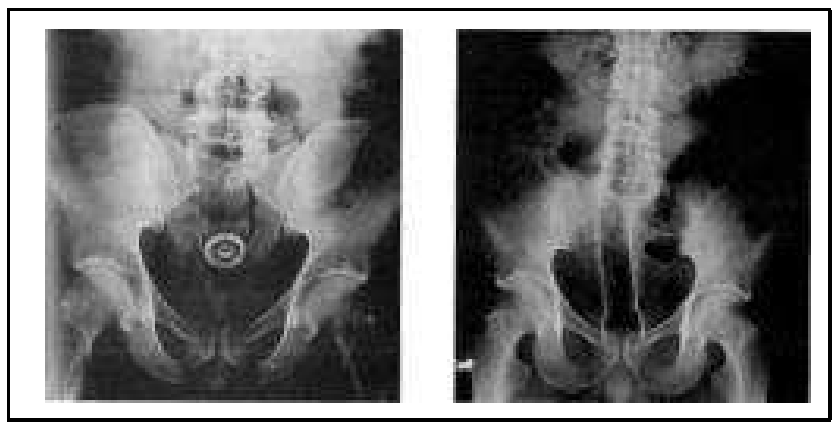

| | |
|---|---|
| $\mathsf{xray}_{hide}$ | $\mathsf{xray\!-\!old}_{hide} \succ \mathsf{xray\!-\!old}_{plain}$ |
| $\neg(\mathsf{xray}_{hide})$ | $\mathsf{xray\!-\!old}_{plain} \succ \mathsf{xray\!-\!old}_{hide}$ |

- $\mathsf{Notes\text{-}old}$ [Textual notes, 2001] can be either presented ($\mathsf{notes\!-\!old}_{plain}$), or omitted all together ($\mathsf{notes\!-\!old}_{hide}$). The preference over the presentation options of $\mathsf{Notes\text{-}old}$ depends on the presentation of $\mathsf{X\text{-}ray\text{-}old}$:

| | |
|---|---|
| $\mathsf{xray\!-\!old}_{hide}$ | $\mathsf{notes\!-\!old}_{plain} \succ \mathsf{notes\!-\!old}_{hide}$ |
| $\mathsf{xray\!-\!old}_{plain}$ | $\mathsf{notes\!-\!old}_{hide} \succ \mathsf{notes\!-\!old}_{plain}$ |

□

At the beginning of a viewing session, the initial presentation of the document, depicted in Figure 8(a), is determined using the forward sweep procedure with $\mathbf{Z} = \emptyset$: each component is set to its preferred presentation given the presentation of its immediate parents in the CP-net. For example, $\mathsf{CT\text{-}image}$ is hidden, since it is the most preferred option for this component. Subsequently, the $\mathsf{X\text{-}ray}$ image is presented segmented, since $\mathsf{CT\text{-}image}$ is not presented, and, in turn, the electromyography $\mathsf{Graph}$ is presented because of the above decision on the presentation options for $\mathsf{CT\text{-}image}$ and $\mathsf{X\text{-}ray}$. Suppose that the viewer chooses to look at the right-top part of the CT-image.[6] In terms of the forward sweep procedure, this choice sets $\mathbf{Z} = \{\mathsf{CT{-}image}\}$, and $\mathbf{z} = \{\mathsf{ct}_{rt}\}$. The result of the forward sweep procedure appears in Figure 8(b); here and in what follows, the shaded nodes stand for the evidence-constrained variables $\mathbf{Z}$. Now, the X-ray image is presented without segmentation because of a zoom-in on the right-top part of CT-image, and Notes are summarized since both the electromyography Graph is presented, and CT-image is not hidden.

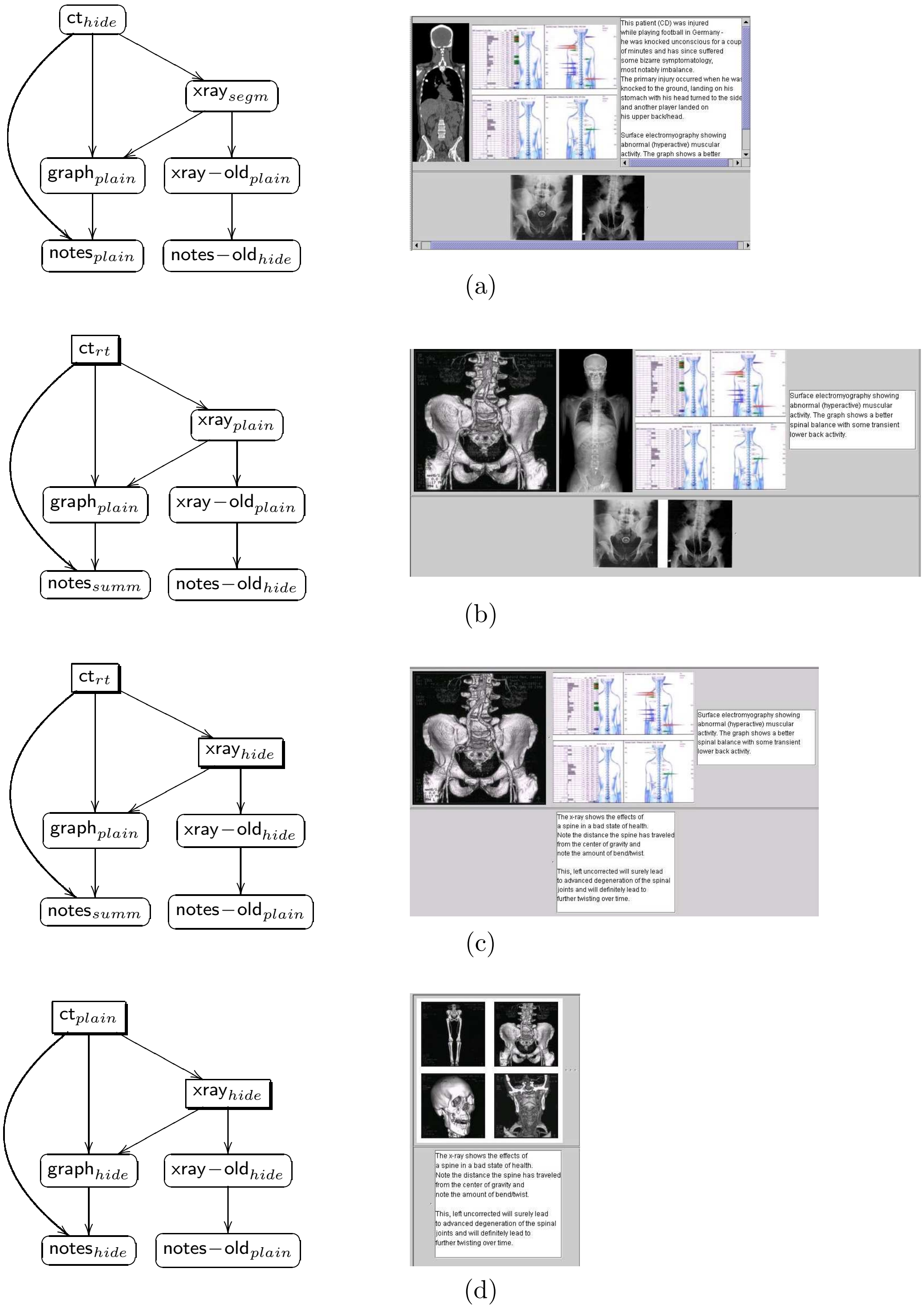

Figure 8: Document presentations after various user choices.

Suppose that the viewer consequently chooses to hide the X-ray image. If the number of recent viewer choices taken to constrain the presentation is greater than one, then this choice will set $\mathbf{Z} = \{\mathsf{CT{-}image}, \mathsf{X{-}ray}\}$, and $\mathbf{z} = \{\mathsf{ct}_{rt}, \mathsf{xray}_{hide}\}$. The result of the forward sweep procedure appears in Figure 8(c). If consequently the viewer chooses to see the whole CT-image, then $\mathbf{z} = \{\mathsf{ct}_{plain}, \mathsf{xray}_{hide}\}$, and the updated presentation is shown in Figure 8(d).

## 4. Comparing Outcomes

Outcome optimization is not the only task that should be supported by a preference representation model. Another basic query with respect to such a model is *preferential comparison* between outcomes. Two outcomes $o$ and $o'$ can stand in one of three possible relations with respect to a CP-net $N$: either $N \models o \succ o'$; or $N \models o' \succ o$; or $N \not\models o \succ o'$ and $N \not\models o' \succ o$.[7] The third case, specifically, means that the network $N$ does not contain enough information to prove that either outcome is preferred to the other (i.e., there exist preference orderings satisfying $N$ in which $o \succ o'$ and in which $o' \succ o$). There are two distinct ways in which we can compare two outcomes using a CP-net:

1. *Dominance queries* – Given a CP-net $N$ and a pair of outcomes $o$ and $o'$, ask whether $N \models o \succ o'$. If this relation holds, $o$ is preferred $o'$, and we say that *$o$ dominates $o'$* with respect to $N$.

2. *Ordering queries* – Given a CP-net $N$ and a pair of outcomes $o$ and $o'$, ask if $N \not\models o' \succ o$. If this relation holds, there exists a preference ordering consistent with $N$ in which $o \succ o'$. In other words, it is consistent with the knowledge expressed by $N$ to order $o$ "over" $o'$ (i.e., assert that $o$ is preferred to $o'$). In such a case we say $o$ is *consistently orderable over $o'$* with respect to $N$.

Ordering queries are clearly weaker than dominance queries. Indeed, if $N \models o \succ o'$, then $N \not\models o' \succ o$. But it may be the case that $N \not\models o' \succ o$ even though $N \not\models o \succ o'$. While dominance queries are typically more useful, ordering queries are sufficient in many applications where one may be satisfied knowing only that outcome $o$ can be consistently ordered over $o'$. For example, consider a set of products that a human or automated seller would like to present to a customer in some non-increasing order of customer preference. There seems to be no reason to use the strong dominance relation to sort such products. In some applications, dominance queries cannot be replaced by ordering queries. For instance, dominance queries were shown to be an integral part of constraint-based preferential optimization in CP-nets (Boutilier, Brafman, Geib, & Poole, 1997).

6. A document explorer (which is a part of the viewing tool) is not illustrated here in order to make the snapshots smaller.
7. Recall that, for the time being, we do not consider CP-nets with indifference in CPTs; hence two outcomes cannot be proven equally preferred.

We begin by showing that ordering queries with respect to acyclic CP-nets can be answered in time linear in the number of variables. In addition, we show that a set of outcomes can be sorted in a consistent non-increasing order with respect to an acyclic CP-net using ordering queries only. We then provide a complexity analysis of dominance queries. First, we introduce and study a particular form of reasoning, namely search for flipping sequences, that can be used to answer dominance queries. Using this technique, and focusing on binary-valued CP-nets, we show connections between the structure of the CP-net graph and the worst-case complexity of dominance queries. We discuss dominance queries in more detail in Section 5, where we present some general search techniques for flipping sequences.

### 4.1 Ordering Queries Are Easy

Here we show that ordering queries with respect to acyclic (not necessarily binary-valued) CP-nets can be answered in time linear in the number of variables. The corresponding algorithm exploits the graphical structure of the model. Likewise, we show that with acyclic CP-nets, we can construct a non-increasing ordering over outcomes, consistent with a CP-net, using only ordering queries.

**Corollary 4** *Let $N$ be an acyclic CP-net, and $o, o'$ be a pair of outcomes. If there exists a variable $X$ in $N$, such that:*

1. *$o$ and $o'$ assign the same values to all ancestors of $X$ in $N$, and*
2. *given the assignment provided by $o$ (and $o'$) to $Pa(X)$, $o$ assigns a more preferred value to $X$ than that assigned by $o'$*

*then $N \not\models o' \succ o$.*

**Proof:** The construction in Theorem 1 provides a preference ordering satisfying $N$ such that $o \succ o'$. Thus $o' \succ o$ is not true in all models of $N$, and is not a consequence of $N$. □

Corollary 4 presents a condition which is sufficient but not necessary for the truth of the ordering query $N \not\models o' \succ o$. For instance, consider Example 2, and let $o = M_{mc} \wedge S_v \wedge W_w$ and $o' = M_{fc} \wedge S_v \wedge W_r$. These two outcomes are incomparable according to the CP-network (i.e., neither can be proven to be preferred to the other), but $o \not\succ o'$ cannot be deduced using the conditions of Corollary 4, because $M$ is the root variable of this chain CP-net, and $o$ assigns it a more preferred value than that assigned by $o'$.

Despite the fact that the condition provided by Corollary 4 for $N \not\models o' \succ o$ is not necessary for consistent orderability, we can show that it is sufficient to provide a consistent ordering of any pair of outcomes.

**Theorem 5** *Given an acyclic CP-net $N$, and two outcomes $o$ and $o'$ over the variables of $N$, the complexity of determining truth of at least one of the ordering queries, $N \not\models o' \succ o$ or $N \not\models o \succ o'$, is $O(n)$.*

**Proof:** For any variable $X_i$, let $Pa(X_i) = \mathbf{U}$ in $N$ and $\mathbf{u}$ and $\mathbf{u}$ denote the assignment to $\mathbf{U}$ made by outcomes $o$ and $o'$, respectively. All variables $X_i$, such that $o$ and $o'$ assign different values to $X_i$ but the same values to all ancestors of $X_i$ in $N$, can be identified in $O(n)$ time by a top-down traversal of $N$. (Note that $\mathbf{u} = \mathbf{u}'$ for all such $X_i$). If for all such $X_i$ we have that $o[X_i] \succ^i_{\mathbf{u}} o'[X_i]$, then using Corollary 4 we can deduce that $N \not\models o' \succ o$. Otherwise, there exist two variables of this type, $X_i$ and $X_j$, for which $o[X_i] \succ^i_{\mathbf{u}} o'[X_i]$ and $o[X_j] \prec^i_{\mathbf{u}} o'[X_j]$; in this case, Corollary 4 implies that both $N \not\models o' \succ o$ and $N \not\models o \succ o'$ □

Corollary 4 provides an effective algorithm for answering ordering queries; however, its computational efficiency comes at a price: it is sound—if the algorithm says that $o$ is consistently orderable over $o'$, then indeed, $N \not\models o' \succ o$; but it is incomplete—if it provides a negative response to query $N \not\models o' \succ o$, it still may be the case that $N \not\models o' \succ o$. Theorem 5 provides an effective algorithm that is sound, and "partially complete" in the sense that it will return a positive answer for at least one of $N \not\models o' \succ o$ or $N \not\models o \succ o'$. In other words, it will allow us to determine that at least one outcome can be consistently ordered over the other.

Though the incompleteness of the algorithm for single ordering queries is problematic, the partial completeness of the algorithm for paired queries is sufficient to allow one to find a consistent ordering of all outcomes in polynomial time, at least in the case of an acyclic CP-net. We first introduce some notation. We write $N \vdash_{oq} o \gg o'$ to represent that the algorithm for paired ordering queries tells us that $N \not\models o' \succ o$ holds (i.e., $o$ is consistently orderable over $o'$) but $N \not\models o' \succ o$ does not (i.e., $o'$ is not orderable over $o$). When $N \vdash_{oq} o \gg o'$, we can be assured that $o$ is indeed orderable over $o'$; but due to the incompleteness of the algorithm, we cannot be sure that $o'$ is not orderable over $o$. We write $N \vdash_{oq} o \simeq o'$ to denote that the algorithm returns a positive response for both ordering queries (i.e., it tells us that both outcomes are consistently orderable over the other). The soundness of the algorithm ensures that both outcomes can indeed be consistently preferred in this case. Note that partial completeness ensures that either $N \vdash_{oq} o \gg o'$, $N \vdash_{oq} o' \gg o$, or $N \vdash_{oq} o \simeq o'$. This will be sufficient to allow us to produce a consistent ordering of any set of outcomes.

**Theorem 6** *Given an acyclic CP-net $N$ over the variable set $\mathbf{V}$, and a set of outcomes $o_1, \ldots, o_m$ over $\mathbf{V}$, ordering these outcomes consistently with $N$ can be done using ordering queries only.*

**Proof:** Define two binary relations over outcomes: $o \gg o'$ iff $N \vdash_{oq} o \gg o'$ and $o \simeq o'$ iff $N \vdash_{oq} o \simeq o'$. We first show that the transitive closure of the relation $\gg$ is asymmetric. Assume to the contrary that there exists a set of outcomes $o_1, \ldots, o_k$ such that:

$$o_1 \gg o_2 \gg \cdots \gg o_k \gg o_1 \tag{1}$$

For $1 \leq i \leq k$, let $V(o_i)$ be the set of all variables $X$ such that the value assigned to $X$ by $o_i$ can be improved given the assignment provided by $o_i$ to $Pa(X)$. Let $N_i$ be the subgraph of $N$ consisting of those variables in $V(o_i)$ and their descendants in $N$. Observe that Corollary 4 implies $N_i \subseteq N_{i+1}$ for $1 \leq i < k$, and $N_k \subseteq N_1$. To see this, notice that if, for some $i$, we have $N_i \not\subseteq N_{i+1}$, then there exists a variable $X$ such that: (i) all ancestors of $X$ are assigned their most preferred values by both $o_i$ and $o_{i+1}$; and (ii) given $o_i[Pa(X)] = o_{i+1}[Pa(X)]$, $X$ is assigned its most preferred value by $o_{i+1}$ and one of its less preferred values by $o_i$.

However, in this case, an ordering query will determine $N \not\models o_i \succ o_{i+1}$, which contradicts our assumption that $o_i \gg o_{i+1}$.

If one of the graph containment relations $N_i \subseteq N_{i+1}$ is strict, the initial assumption (1) is trivially contradicted. Therefore, we are left with the case of:

$$N_1 = N_2 = \cdots = N_k = N'$$

Recalling that $N'$ is acyclic, consider a variable $X_j \in N'$ that has no ancestors in $N'$. Let $\mathbf{U} = Pa(X_j)$ be the parents of $X_j$ in the original network $N$ (note that $\mathbf{U} \cap N' = \emptyset$). By construction of the $N_i$ we have:

$$o_1[\mathbf{U}] = o_2[\mathbf{U}] = \cdots = o_k[\mathbf{U}] = \mathbf{u}$$

This must be the case since all the ancestors of $X_j$ are assigned to their unique optimal assignment (of which $\mathbf{u}$ is a part) since none of these variables is improvable. This entails

$$o_1[X_j] \succ^j_{\mathbf{u}} o_2[X_j] \succ^j_{\mathbf{u}} \cdots \succ^j_{\mathbf{u}} o_k[X_j] \succ^j_{\mathbf{u}} o_1[X_j],$$

which is inconsistent with the definition of a CP-net.

We exploit the asymmetric nature of the relation $\gg$ as follows. If $N \models o \succ o'$, then we must have $o \gg o'$. Therefore, the relation $\succ_N$ representing the induced preference graph of $N$ is a subset of $\gg$. Thus any total ordering of $o_1, \ldots, o_m$ consistent with $\gg$ will be consistent with $\succ_N$. $\Box$

An immediate consequence of Theorems 5 and 6 is that, given a set of $m$ outcomes and a CP-net $N$, the complexity of ordering these outcomes consistently with the preference graph induced by $N$ is $O(nm^2)$ (i.e., the cost of comparing every pair of outcomes and ordering them accordingly).

## 4.2 Dominance Queries and Flipping Sequences

The *ceteris paribus* semantics of CP-nets allows one to directly use information in the CPT of a variable $X$ to alter or *flip* the value of $X$ within an outcome to directly obtain an improved (preferred) or worsened (dispreferred) outcome. A *sequence of improving flips* from one outcome to another provides a proof that one outcome is preferred, or dispreferred, to another in all rankings satisfying the network. Before defining this notion more precisely, we illustrate the intuitions with an example.

**Example 6** Consider again the CP-net from Figure 4. The following are the only two rankings that satisfy this network:

$$abc \succ ab\bar{c} \succ a\bar{b}\bar{c} \succ \overbrace{a\bar{b}c \succ \bar{a}\bar{b}\bar{c}} \succ \bar{a}\bar{b}c \succ \bar{a}bc \succ \bar{a}b\bar{c}$$
$$abc \succ ab\bar{c} \succ a\bar{b}\bar{c} \succ \underbrace{\bar{a}\bar{b}\bar{c} \succ a\bar{b}c} \succ \bar{a}\bar{b}c \succ \bar{a}bc \succ \bar{a}b\bar{c}$$

Thus, the only two outcomes not totally ordered are $\overline{a}\overline{b}\overline{c}$ and $a\overline{b}c$. Notice that if we remove either $\overline{a}\overline{b}\overline{c}$ or $a\overline{b}c$ from each of these chains of outcomes, we can move from one outcome to the next in the chain by *flipping* the value of exactly one variable according to the preference information in its CPT given the instantiation of its parents. For example, to move from the first outcome in these sequences ($abc$) to the second ($ab\overline{c}$), we use the fact that $c \succ \overline{c}$ given $b$ to "prove" that the second outcome is dispreferred to the first; that is we *flip* $C$ to a less preferred value given the instantiation $b$ of its parent $B$. Conversely, we can move backwards through this sequence by flipping $\overline{c}$ in the second outcome to $c$, thereby obtaining the more preferred first outcome.

Recall that Corollary 4 demonstrates that violating the preference constraints for a parent variable is less preferred than violating the preference constraints for any of its children. This "greater importance" of parent variables is implicit in the *ceteris paribus* semantics. Now consider the two outcomes $\overline{a}\overline{b}\overline{c}$ and $a\bar{b}c$, which are unordered by the CP-net in Figure 4. The outcome $\bar{a}b\bar{c}$ violates the preference over the values of $A$, while the outcome $a\bar{b}c$ violates the preferences over the values of $B$ and $C$; and $A$ is an ancestor of both $B$ and $C$. The semantics of CP-nets does not specify which of these outcomes is preferred—intuitively, though the preference for $A$ has higher priority than $B$ or $C$, two or more violations of lower priority preferences may not be preferred to the violation of a single higher priority preference. □

For any two outcomes $o$ and $o'$, every improving flipping sequence from $o'$ to $o$ uniquely corresponds to some directed path from the node $o'$ to the node $o$ in the preference graph induced by the CP-net. For instance, consider the CP-net in Figure 9(a), which is exactly the network of the "Evening Dress" example (Example 3), with simpler variable names. There are four alternative flipping sequences from the outcome $\bar{a}\bar{b}c$ to the outcome $abc$, corresponding to the four paths between these outcomes in the induced preference graph, depicted in Figure 9(b):

$$\begin{aligned}
&\bar{a}\bar{b}c \rightarrow a\bar{b}c \rightarrow abc \\
&\bar{a}\bar{b}c \rightarrow a\bar{b}c \rightarrow a\bar{b}\bar{c} \rightarrow ab\bar{c} \rightarrow abc \\
&\bar{a}\bar{b}c \rightarrow \bar{a}bc \rightarrow abc \\
&\bar{a}\bar{b}c \rightarrow \bar{a}bc \rightarrow \bar{a}b\bar{c} \rightarrow ab\bar{c} \rightarrow abc
\end{aligned}$$

Therefore, $abc \succ \bar{a}\bar{b}c$ is a consequence of this CP-net. In contrast, there is no flipping sequence (directed path) from $\bar{a}bc$ to $a\bar{b}\bar{c}$, hence these two outcomes are incomparable.

These examples suggest that the construction of such a flipping sequence can be used to prove dominance.

**Definition 4** Let $N$ be a CP-net over variables $\mathbf{V}$, with $X_i \in \mathbf{V}$, $\mathbf{U}$ the parents of $X_i$, and $\mathbf{Y} = \mathbf{V} - (\mathbf{U} \cup \{X_i\})$. Let $\mathbf{u}x\mathbf{y} \in \mathit{Asst}(\mathbf{V})$ be any outcome, where $x \in \mathit{Dom}(X_i)$, $\mathbf{u} \in \mathit{Asst}(\mathbf{U})$, and $\mathbf{y} \in \mathit{Asst}(\mathbf{Y})$. An *improving flip* of outcome $\mathbf{u}x\mathbf{y}$ with respect to variable $X_i$ is any outcome $\mathbf{u}x'\mathbf{y}$ such that $x' \succ^i_{\mathbf{u}} x$. (Note that an improving flip w.r.t. $X_i$ does not exist if $x$ is the most preferred value of $X_i$ given $\mathbf{u}$.)

An *improving flipping sequence* with respect to $N$ is any sequence of outcomes $o_1, \ldots, o_k$ such that, for each $i < k$, $o_{i+1}$ is an improving flip of $o_i$ with respect to some variable. An improving flipping sequence from an outcome $o$ to an outcome $o'$ is any improving sequence $o_1, \ldots, o_k$ with $o_1 = o$ and $o_k = o'$.

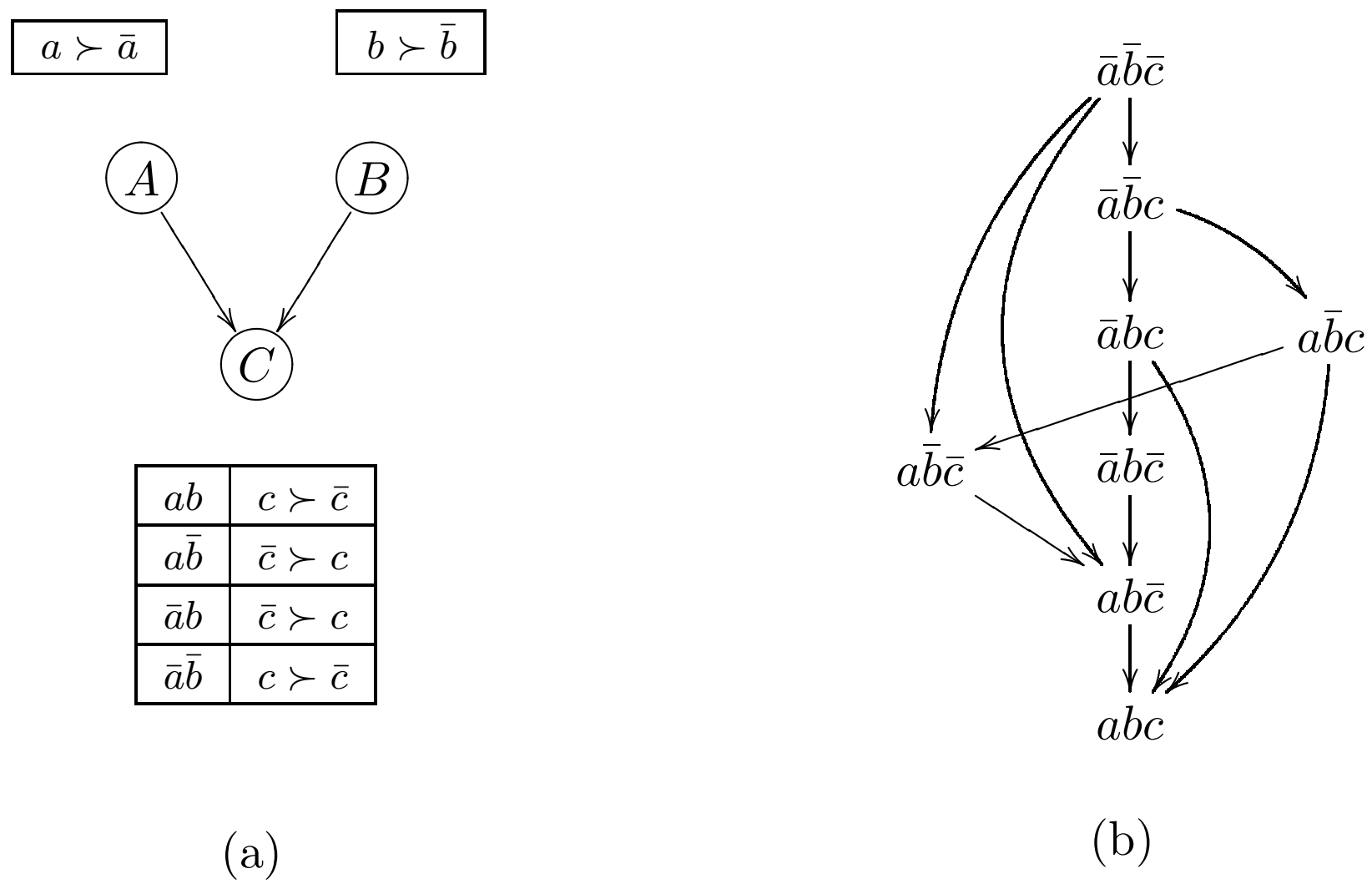

Figure 9: “Evening Dress” CP-net from Example 2 with simpler names.

One can define *worsening flips* and *worsening flipping sequences* in an entirely analogous way. Obviously, any worsening flipping sequence is the reverse of an improving flipping sequence, and vice versa.

There are two important things to notice about the examples above. First, an improving (or worsening) flipping sequence can be used to show that one outcome is better than another. Second, preference violations are worse (i.e., have a larger negative impact on the preference of an outcome) the higher up they are in the network, although we cannot compare always two (or more) lower level violations to violation of a single ancestor constraint. These observations underlie the inference algorithms below.

**Theorem 7 (soundness)** *If there is an improving flipping sequence for CP-net $N$ from outcome $o$ to $o'$, then $N \models o' \succ o$.*

**Proof:** If there is an improving flip from outcome $o_1$ to another outcome $o_2$ then $N \models o_2 \succ o_1$ by the definition of $\models$. The theorem follows from the transitivity of preferential entailment (Lemma 2). $\square$

**Theorem 8 (completeness)** *If $N$ is an acyclic CP-net and there is no improving flipping sequence for $N$ from outcome $o$ to $o'$, then $N \not\models o' \succ o$.*

**Proof:** Let $G$ be a graph whose nodes are all outcomes (i.e., complete assignments to the variable set $\mathbf{V}$), with a directed edge from $o_1$ to $o_2$ iff there is an improving flip of $o_1$ to $o_2$ sanctioned by network $N$. Clearly, directed paths in $G$ are equivalent to improving flipping sequences with respect to $N$.

Next, we show that any total preference ordering $\succ$ that respects the paths in $G$ (that is, if there is a path from $o_1$ to $o_2$ in $G$, then we have $o_2 \succ o_1$) satisfies network $N$: if $\succ$ does not satisfy $N$, there must exist some variable $X$ with parents $\mathbf{U}$, instantiation $\mathbf{u} \in Asst(\mathbf{U})$, values $x, x' \in Dom(X)$, and instantiation $\mathbf{y}$ of the remaining variables $\mathbf{Y} = \mathbf{V} - (\mathbf{U} \cup \{X\})$, such that:

(a) $\mathbf{u}x\mathbf{y} \succ \mathbf{u}x'\mathbf{y}$;

(b) but $CPT(X)$ dictates that $x' \succ x$ given $\mathbf{u}$.

This is a direct consequence of the definition of satisfaction. However, if $N$ requires that $x' \succ x$ given $\mathbf{u}$, there is a direct flip from $x\mathbf{uy}$ to $x'\mathbf{uy}$, contradicting the fact that $\succ$ extends the graph $G$.

Based on this observation, we can now prove the theorem: If there is no improving flipping sequence from $o$ to $o'$, then there is no directed path in $G$ from $o$ to $o'$. Therefore, there exists a preference ordering $\succ$ respecting the paths in $G$ in which $o \succ o'$. But this preference ordering also satisfies $N$, which implies $N \not\models o' \succ o$. □

### 4.3 Flipping Sequences as Plans

Searching for flipping sequences can be seen as a type of planning problem: given a CP-net $N$, and a variable $X$ with parents $Pa(X)$ in $N$, each row in $CPT(X)$ is a conditional preference statement of the form

$$\mathbf{u} : x_1 \succ x_2 \succ \cdots \succ x_d$$

where $\mathbf{u} \in Asst(Pa(X))$, and $d = |Dom(X)|$. Such a statement can be converted into a set of planning operators for improving the value of $X$. In particular, this conditional preference statement can be converted into a set of $d-1$ planning operators of the form (for $1 < i \leq d$):

**Preconditions:** $\mathbf{u} \wedge x_i$

**Postconditions:**

**Delete list:** $x_i$

**Add list:** $x_{i-1}$

This corresponds to the action of improving $x_i$ to $x_{i-1}$ in the context of $\mathbf{u}$. (An "inverse" set of operators would be created for worsening sequences).

Given a query $N \models \mathbf{x} \succ \mathbf{y}$, we treat $\mathbf{y}$ as the start state and $\mathbf{x}$ as the goal state of a planning problem. It is readily apparent that the query is a consequence of the CP-net if and only if there is a plan for the associated planning problem, since a plan corresponds to a flipping sequence.

The planning problem over multi-valued variables with discrete, finite domains is known to be PSPACE-complete (Bäckström & Nebel, 1995), and it remains PSPACE-complete under the assumption that all the variables are binary (Bylander, 1994) (i.e., planning problems in STRIPS formalism with negative effects). However, this upper bound is not very informative with respect to dominance queries, since the planning problems generated from them will generally look quite different in form from standard AI planning problems, as there are many more actions, and each action is directed toward achieving a particular proposition and requires very specific preconditions. Thus dominance queries with respect to binary-valued CP-nets correspond to a specific class of STRIPS planning problems, the complexity of which was recently analyzed by Domshlak and Brafman (2002b, 2003). We now explain this relationship.

First, we divide the preconditions of every operator in a planning problem into two types: *prevailing* conditions, which are variable values that are required prior to the execution of the operator and are not affected by the operator, and *preconditions*, which *are* affected by the operator. Second, we introduce the notion of a *causal graph* (Knoblock, 1994), a directed graph whose nodes stand for the problem variables. An edge $(X, Y)$ appears in the causal graph if and only if some operator that changes the value of $Y$ has a prevailing condition involving $X$.

The complexity analysis of Brafman and Domshlak (2003) addresses planning problems with binary variables, unary operators (i.e., operators that affect only a single variable), and acyclic causal graphs. In the planning problem generated for a dominance query with respect to a binary-valued CP-net, we have:

1. all the operators are unary, because each flip improves the value of a single variable; and

2. the causal graph of the problem is exactly the graph of the CP-net, since the values of $Pa(X)$ required by a value flip of $X$ are exactly the prevailing conditions of the corresponding planning operator.

Therefore, in our computational analysis of dominance queries for binary-valued acyclic CP-nets we can use some of the results and techniques of Brafman and Domshlak (2003).

### 4.4 Complexity of Dominance Queries for Binary-valued, Acyclic CP-nets

In this section we analyze the complexity of dominance testing with respect to binary-valued CP-nets, showing a connection between the structure of the CP-net graph and the worst-case complexity of dominance queries. In particular, we show that:

- When a binary-valued CP-net forms a directed tree, the complexity of dominance testing is quadratic in the number of variables.

- When a binary-valued CP-net forms a polytree (i.e., the induced undirected graph is acyclic), dominance testing is polynomial in the size of the CP-net description.

- When a binary-valued CP-net is directed-path singly connected (i.e., there is at most one directed path between any pair of nodes), dominance testing is NP-complete. The problem remains hard even if node in-degree in the network is bounded by a low constant.

- Dominance testing for binary-valued CP-nets remains NP-complete if the number of alternative paths between any pair of nodes in a CP-net is polynomially bounded.

The exact complexity of dominance testing in multiply connected, binary-valued, acyclic CP-nets remains an open problem—at this stage it is not clear whether this problem is in NP or harder.

In what follows, we make the assumption that the number of parents for each variable (i.e., node in-degree in the CP-net) is bounded by some constant. This assumption is justified as the CPTs are part of the problem description, and the size of a $CPT(X)$ is exponential in $|Pa(X)|$.

### 4.4.1 Some General Properties

We start with some notation and two useful lemmas. First, given a CP-net $N$ and a pair of outcomes $o, o'$ with respect to $N$, an improving flipping sequence $F$ from $o'$ to $o$ will be called *irreducible* if any subsequence $F'$ of $F$ obtained by deletion of any entries except the endpoints $o, o'$ of $F$ is not an improving flipping sequence.[8]

Given a CP-net $N$, let $\mathcal{F}$ be the set of *all* irreducible improving flipping sequences among outcomes. We denote by $\mathsf{MaxFlip}(X_i)$ the maximal number of times that a variable $X_i$ changes its value in any flipping sequence in $\mathcal{F}$. Formally, let $\mathsf{Flip}(F, X_i)$ be the number of value flips of $X_i$ in the flipping sequence $F$. Then,

$$\mathsf{MaxFlip}(X_i) = \max_{F \in \mathcal{F}} \{\mathsf{Flip}(F, X_i)\}$$

Lemma 9 below formalizes our first observation about irreducible flipping sequences with respect to binary-valued CP-nets.

**Lemma 9** *For every variable $X_i$ in a binary-valued CP-net $N$, we have:*

$$\mathit{MaxFlip}(X_i) \leq \; 1 \; + \sum_{X_j : X_i \in Pa(X_j)} \mathit{MaxFlip}(X_j)$$

**Proof:** Let $F$ be an irreducible flipping sequence with respect to $N$ (from some outcome $o'$ to some outcome $o$), such that $\mathsf{MaxFlip}(X_i) = \mathsf{Flip}(F, X_i)$. Consider the subsequence $F' = f_1, f_2, \ldots, f_k \subseteq F$ that consists of all value flips of the children of $X_i$ in $N$. Observe that: (i) every $f_l \in F'$ *requires* $X_i$ to take one of its two possible values; and (ii) no value flip in $F - F'$ depends on the value of $X_i$.

Now, for $1 \leq l < k$, if $f_l$ and $f_{l+1}$ require the same value of $X_i$, then there are no value flips of $X_i$ in $F$ between $f_l$ and $f_{l+1}$: If there are such flips, they are simply redundant, and this contradicts our assumption that $F$ is irreducible. (Recall that $f_l$ and $f_{l+1}$ are adjacent in $F'$, but may be separated by several flips in the original sequence $F$.) Alternatively, if $f_l$ and $f_{l+1}$ require different values of $X_i$, due to the irreducibility of $F$, there is exactly one value flip of $X_i$ in $F$ between $f_l$ and $f_{l+1}$. Similarly we can show that there is at most one value flip of $X_i$ in $F$ before $f_1$, and there is at most one value flip of $X_i$ in $F$ after $f_k$. The latter flip is necessary when $f_k$ requires $X_i$ to take the value $\neg o[X_i]$, thus, after "supporting" the immediate successors, $X_i$ still should flip its value once, in order to obtain the value required by $o$.

The above implies that:

$$\mathsf{MaxFlip}(X_i) \; = \; \mathsf{Flip}(F, X_i) \; \leq \; 1 + \sum_{X_j : X_i \in Pa(X_j)} \mathsf{Flip}(F, X_j)$$

8. Note that removing any proper initial or final subsequence of $F$ results in a valid flipping sequence. We refer here to the deletion of arbitrary elements from the sequence, *excluding* the endpoints.

and thus, by definition of MaxFlip, we have:

$$\mathsf{MaxFlip}(X_i) \;\leq\; 1 + \sum_{X_j : X_i \in Pa(X_j)} \mathsf{MaxFlip}(X_j)$$

□

Adopting the terminology of Domshlak and Shimony (2003) and Shimony and Domshlak (2002), a directed acyclic graph $G$ is *directed-path singly connected* if, for every pair of nodes $s, t \in G$, there is at most one directed path from $s$ to $t$. Using Lemma 9, we can prove that if a binary-valued CP-net forms a directed-path singly connected DAG then, for every variable $X_i$, $\mathsf{MaxFlip}(X_i)$ can be bounded by $n$ (the number of variables).

**Lemma 10** *If a binary-valued CP-net $N$ forms a directed-path singly connected DAG, then, for every variable $X_i \in N$, we have:*

$$\mathit{MaxFlip}(X_i) \leq n$$

*where $n$ is the number of variables in $N$.*

**Proof:** In what follows, denote $\mathsf{MaxFlip}(X_i)$ in $N$ as $\mathsf{MaxFlip}_N(X_i)$. The proof is by induction on $n$. For $n = 1$ it is obvious that $\mathsf{MaxFlip}(X_1) \leq 1$. Assume that when a binary-valued, directed-path singly connected CP-net $N$ consists of $n-1$ variables, then, for every $X_i \in N$,

$$\mathsf{MaxFlip}_N(X_i) \leq n - 1$$

Let $N'$ be some binary-valued, directed-path singly connected CP-net over $n$ variables. Without loss of generality, let the variables $\{X_1, \dots X_n\}$ of $N'$ be topologically ordered based on the (acyclic) graph of $N'$. Clearly, $X_n$ is a leaf node, and we will denote by $N$ the CP-net obtained by removing $X_n$ from $N'$. From Lemma 9, we have:

$$\mathsf{MaxFlip}_{N'}(X_n) \leq 1$$

Observe that there are no directed paths between any predecessors of $X_n$ in $N'$, since $N'$ is assumed to be directed-path singly connected. Therefore, by Lemma 9, for each parent $X_i$ of $X_n$ in $N'$, we have:

$$\mathsf{MaxFlip}_{N'}(X_i) \leq \mathsf{MaxFlip}_N(X_i) + \mathsf{MaxFlip}_{N'}(X_n)$$

and thus:

$$\mathsf{MaxFlip}_{N'}(X_i) \leq \mathsf{MaxFlip}_N(X_i) + 1$$

Generally, since $N$ is directed-path singly connected, for each variable $X_i \in N'$,

$$\mathsf{MaxFlip}_{N'}(X_i) \leq \begin{cases} \mathsf{MaxFlip}_N(X_i) + 1, & \text{if there is a path from } X_i \text{ to } X_n \\ \mathsf{MaxFlip}_N(X_i), & \text{otherwise} \end{cases}$$

and thus, for each $X_i \in N'$, we have:

$$\mathsf{MaxFlip}_{N'}(X_i) \leq n$$

□

```
TreeDT ([[N ⊨ o ≻ o′]])
Initialize the variables in V to outcome o′.
Loop:
   Iteratively remove all leaf variables from V that have assigned to
      them their values in o.
   If V = ∅, then return yes.
   Find a variable X s.t. its value can be improved, and no value of its
      descendants in N can be improved, given the current assignment to V.
   If no such variable was found, then return no.
      Otherwise, change the value of X.
```

Figure 10: Flipping sequence search algorithm for binary-valued, tree-structured CP-nets

#### 4.4.2 Tree-structured CP-nets

We start by presenting a flipping sequence search algorithm for the class of binary-valued, tree-structured CP-nets, and prove its correctness. Then we show that the time complexity of this algorithm is $O(n^2)$, and show that this is actually a lower bound for flipping-sequence search over binary-valued, tree-structured CP-nets.

Figure 10 presents our TreeDT algorithm for binary-valued, tree-structured CP-nets. Informally, TreeDT starts by initializing all the variables in $\mathbf{V}$ to their values in the (purported) less preferred outcome $o'$, and continues with an incremental, *bottom-up* conversion of this initial assignment to the assignment induced by the (purported) more preferred outcome $o$. Each step starts by iteratively removing from $N$ all leaf variables (i.e., its maximal fringe or canopy) that are already consistent with $o$. If at some iteration this step removes all variables in $N$, then all variables are assigned to their values in $o$, and thus the required improving flipping sequence from $o'$ to $o$ has been generated.

Otherwise, let $N'$ stand for the updated $N$ (with nodes removed). A node $X \in N'$ is a *candidate* variable if: (i) the value of $X$ can be flipped; and (ii) no descendant of $X$ in $N'$ can have its value flipped, given the current assignment to the variables of $N'$. We then flip the value of an *arbitrary* candidate variable (if one exists) and repeat with node removal; or we report that there is no improving flipping sequence from $o'$ to $o$ if there are no candidate variables.

The TreeDT algorithm is deterministic and backtrack-free. Below we show that TreeDT is complete for binary-valued, tree-structured CP-nets, and generates only irreducible flipping sequences—thus the time complexity of TreeDT is $O(n^2)$. The fact that it generates irreducible flipping sequences ensures its soundness (since by generating only valid flipping sequences it can only provide correct positive answers to dominance queries).

**Theorem 11** *The algorithm* TreeDT *is sound and complete for binary-valued, tree-structured CP-nets.*

**Proof:** Consider an execution of TreeDT on a dominance query $N \models o \succ o'$ with respect to a binary-valued, tree-structured CP-net $N$.

First, suppose we iteratively remove from the tree any leaf variables that have as values those required by the target outcome $o$. It is easy to see that this does not affect the completeness of the algorithm: because $N$ is acyclic, the variables in the fringe are not the ancestors of any other variables. Hence, the value of the variables in the fringe does not influence our ability to flip the values of any other variables (hence it does not remove any improving flipping sequence from consideration).

Second, consider a variable $X$ such that, after iteratively removing variables as above, its value can be improved, yet none of its descendents in $N$ can be improved, given the current assignment $\mathbf{v}$ to $\mathbf{V}$. If $X$ is a leaf node, then changing its value does not influence our ability to flip values of any other variables. In addition, the current value $\mathbf{v}[X]$ of $X$ is different from $o[X]$, otherwise $X$ would have been part of the removed fringe. Therefore, the (improving) change of $X$'s value at this point is necessary in any improving flipping sequence. Alternatively, suppose that $X$ is not a leaf node. Since the leaf nodes in the subtree of $N$ rooted at $X$ were not a part of the removed fringe, (at least) their values remain to be changed. Because $N$ is a tree, $X$ completely separates its descendents in $N$ from all other variables; so no improving flip in the subtree of $X$ will be possible until we change the value of $X$. Hence the value of $X$ must be changed in any flipping sequence from this point before the value of any descendent of $X$.

What remains to be shown is that when there are several candidate variables that can be flipped, it does not matter which one we flip first. If there is more than one candidate variable, each one of them will have to be flipped at some point—each such flip is necessary for flipping the children of the corresponding variable. However, any changes made to one of these candidates or below it has no affect on the other candidates or their descendants. Thus, the evaluation order of the candidate flips is irrelevant, and cannot prevent us from finding a flipping sequence if one exists.

Thus the algorithm is complete. The soundness of the algorithm should be clear from the proof as well. The flip generated at each step of the algorithm is a valid improving flip given the current outcome $\mathbf{v}$. □

**Theorem 12** *The time complexity of the flipping-sequence search over binary-valued, tree-structured CP-nets is $O(n^2)$, where $n$ is the number of variables in the CP-net.*

**Proof:** Since the algorithm TreeDT is backtrack-free, the only thing that remains to be shown in addition to Lemma 10 and Theorem 11 is that (on binary-valued, tree-structured CP-nets) TreeDT generates only irreducible flipping sequences. However, the proof of this subclaim is straightforward since we have already shown that:

1. TreeDT flips the value of each variable $X$ either to achieve the value of $X$ in the (purported) more preferred outcome of the query, or to allow the required value flips of the descendants of $X$, and

2. The role of the latter flips of $X$ cannot be fulfilled by value flips of any other variables.

Therefore, given an improving flipping sequence $F$ generated by TreeDT, removing any subset of value flips from $F$ makes $F$ either illegal, or its ends are not $(o', o)$, respectively. Hence, any improving flipping sequence generated by the TreeDT algorithm is irreducible. □

Theorem 13 below shows that even when limiting ourselves to *chain* binary-valued CP-nets, there are dominance queries whose minimal flipping sequences have quadratic length. Thus TreeDT is asymptotically optimal.

**Theorem 13** $\Theta(n^2)$ *is a lower bound for the flipping-sequence search over binary-valued tree-structured CP-nets.*

**Proof:** For the proof see Appendix A. The proof is by example, providing a dominance query on a binary-valued, tree CP-net for which the size of a minimal flipping sequence is $\Theta(n^2)$.

### 4.4.3 Polytree CP-nets

DAGs with no cycles in the underlying undirected graphs, also known as *polytrees*, offer a minimal extension of directed trees. Unfortunately, the TreeDT procedure is not complete for polytree CP-nets unless extended with some form of backtracking (even when restricted to binary variables). This is due to the fact that several parents of a given node may each be allowed to have their values flipped, but only one of the choices may lead to the target outcome while the others lead to a dead end. For instance, consider the CP-net $N$ from Figure 9 and the query $N \models a\bar{b}\bar{c} \succ \bar{a}\bar{b}c$. Starting with the outcome $\bar{a}\bar{b}c$, in the first iteration of TreeDT, we have the choice of flipping either $A$ or $B$. If $B$ is chosen, the assignment is changed to $\bar{a}bc$. However, this cannot lead to the target, since there is no way to flip $B$ back to $\bar{b}$. Thus, a dead end is reached. On the other hand, if $A$ is chosen then TreeDT will successfully generate the improving flipping sequence $\bar{a}\bar{b}c \rightarrow a\bar{b}c \rightarrow a\bar{b}\bar{c}$. Thus, an incorrect choice of improving variable may require backtracking.

However, dominance testing for binary-valued, polytree CP-nets remains polynomial time, although the algorithm for its solution is more complicated than TreeDT. An algorithm is adapted from the corresponding algorithm for planning problems with binary variables, unary operators, and polytree causal graphs described by Domshlak and Brafman (2003).

**Theorem 14** *Dominance testing for binary-valued, polytree CP-nets is in* P.

**Proof:** According to the reduction from the CP-net dominance queries to the classical planning problems (see Section 4.3), every dominance query with respect to a binary-valued, polytree CP-net can be compiled into a propositional planning problem with unary operators and polytree causal graph. An algorithm for the latter problem is presented by Brafman and Domshlak (2003), and the time complexity of this algorithm is $O(2^{2\kappa}n^{2\kappa+3})$, where $\kappa$ is the maximal in-degree of the causal graph.

Recall our assumption that the in-degree of the CP-net is bounded by some constant (in this case, $\kappa$). This assumption is justified as the CPTs are part of the problem description, and the size of a $CPT(X)$ is exponential in $|Pa(X)|$. Therefore, the complexity of the algorithm of Brafman and Domshlak (2003) on CP-nets is polynomial in the size the CP-net. □

#### 4.4.4 Intractable dominance queries

While dominance testing for binary-valued, polytree CP-nets is polynomial, we can show that for binary-valued, directed-path singly connected CP-nets, this problem is NP-complete.[9] This can be proved using a CP-net-oriented extension of the proof for the corresponding claim with respect to planning problems (Brafman & Domshlak, 2003). These results also entail that dominance testing for binary-valued CP-nets remains in NP if the number of distinct paths between any pair of nodes in the CP-net is polynomially bounded.

**Theorem 15** *Dominance testing for binary-valued, directed-path singly connected CP-nets is* NP*-complete.*

**Proof:** For the proof see Appendix A.

An immediate extension of directed-path singly connected DAGs are *max-$\delta$-connected* DAGs: a directed graph is max-$\delta$-connected if the number of different directed paths between any two nodes in the graph is bounded by $\delta$.

**Theorem 16** *Dominance testing for binary-valued, max-$\delta$-connected CP-nets, where $\delta$ is polynomially bounded in the size of the CP-net, is* NP*-complete.*

**Proof:** The theorem is immediately entailed by the corresponding result for planning of Brafman and Domshlak (2003); for the proof in terms of CP-nets, see Appendix A.

Theorem 15 implies that dominance testing for binary-valued, acyclic CP-nets is hard. However, the exact complexity of this problem is still an open question—it is not clear whether this problem is in NP or if it is harder. Some preliminary analysis of this problem (Domshlak & Brafman, 2002a) shows a connection between the complexity of the flipping sequence search for binary-valued, acyclic CP-nets and the diameters of some specific graphs. A complementary result with respect to these graphs (Domshlak, 2002b), namely recursively directed hypercubes, shows that dominance queries with respect to binary-valued, acyclic CP-nets with *unbounded* node in-degree may require flipping sequences of size exponential in $n$, where $n$ is the number of variables in the CP-net.

It has been shown that the most general class of planning problems with binary variables, unary operators and acyclic causal graphs is harder than NP (Brafman & Domshlak, 2003). However, this result does not imply that answering dominance queries is harder than NP as well, as we do not know of a reduction from this class of planning problems into CP-nets.

## 5. Search Techniques for Dominance Queries

In the previous section we showed that dominance testing is generally hard even for binary-valued CP-nets, and that tractable algorithms exist only for specific problem classes. However, CP-nets impose a rich structure on preferences that can be exploited by various search strategies and heuristics which often significantly reduce search effort, and allow the effective solution of many problem instances. In this section we discuss the search for flipping

9. While every polytree is directed-path singly connected, the converse is not true.

sequences and several rules that allow significant pruning of the search tree without impacting soundness or completeness of the search procedure. These rules are described in the context of improving flipping sequences, but they can be applied to worsening search *mutatis mutandis.*

Given a CP-net $N$, and an outcome $o$, we define the *improving search tree* of $o$, $T(o)$, as follows: $T(o)$ is rooted at $o$, and the children of every node $o'$ in $T(o)$ are those outcomes that can be reached by a single improving flip from $o'$. It is easy to see that every rooted path in $T(o)$ corresponds to some improving flipping sequence from the outcome $o$ (with respect to $N$), and vice versa.[10] For example, consider the preference graph shown in Figure 11(a), which is induced by the CP-net in Figure 4. Figure 11(b) depicts the improving search tree $T(\bar{a}b\bar{c})$ with respect to this preference graph. Clearly, we can treat every dominance query $N \models o \succ o'$ as searching $T(o')$ for a node associated with the outcome $o$, starting from the root node $o'$. For instance, in the example above, given the dominance query $N \models a\bar{b}\bar{c} \succ \bar{a}b\bar{c}$, any of the dotted paths shown in Figure 11(c) bring us to outcome $a\bar{b}\bar{c}$. However, taking a different direction during the tree traversal would lead to a dead end, requiring backtracking in order to find a suitable flipping sequence. Any generic search algorithm can be used to traverse the improving search tree to find an improving sequence that supports a dominance query.

### 5.1 Suffix Fixing

*Suffix fixing* is a rule that allows certain moves in $T(o)$ to be pruned from the search tree without impacting completeness of the search. Let $N$ be a CP-net over the variables $\mathbf{V} = \{X_1, \ldots, X_n\}$, numbered according to an arbitrary topological ordering consistent with the network $N$. We define an *rth suffix* of an outcome $o \in Asst(\mathbf{V})$, for any $r \geq 1$, to be the subset of the outcome values $o[X_r]o[X_{r+1}] \cdots o[X_n]$. Notice that the $r$th suffix of an outcome depends on the topological ordering of variables used. Finally, we say that the $r$th suffixes of outcomes $o$ and $o'$ *match* iff $o[X_j] = o'[X_j]$ for all $r \leq j \leq n$.

Let $o^*$ be a node in an improving search tree $T(o')$, and let $o$ be the target of the search; in other words, we are attempting to find a flipping sequence that proves $N \models o \succ o'$. The *suffix fixing rule* requires that if the $r$th suffix of $o^*$ and $o$ matches, for some value $r$, no neighbors of $o^*$ can be explored whose $r$th suffix does not also match $o$. This is equivalent to ruling out the exploration of any flipping sequences that destroy the suffix of an outcome that matches the target outcome $o$. The following lemma ensures that pruning such branches of the search tree $T(o')$ when searching for a path to $o$ does not affect completeness.

**Lemma 17** *Let $N$ be a CP-net, and $o$ and $o^*$ be outcomes whose $r$th suffix matches (for any topologically consistent ordering $X_1, \cdots, X_n$ of the variables in $N$). If there is path in $T(o^*)$ from root $o^*$ to $o$, then there is a path from root $o^*$ to $o$ such that every outcome on that path assigns the same values to $X_r, \cdots X_n$ (i.e., with the same suffix match).*

**Proof:** The proof is straightforward: Because $N$ is acyclic, no suffix variable is an ancestor of any non-suffix variables. Hence, the value of the suffix variables does not influence (for better or worse) our ability to flip values of the remaining variables. □

10. The corresponding worsening search tree of $o$ can be defined similarly using the worsening flips.

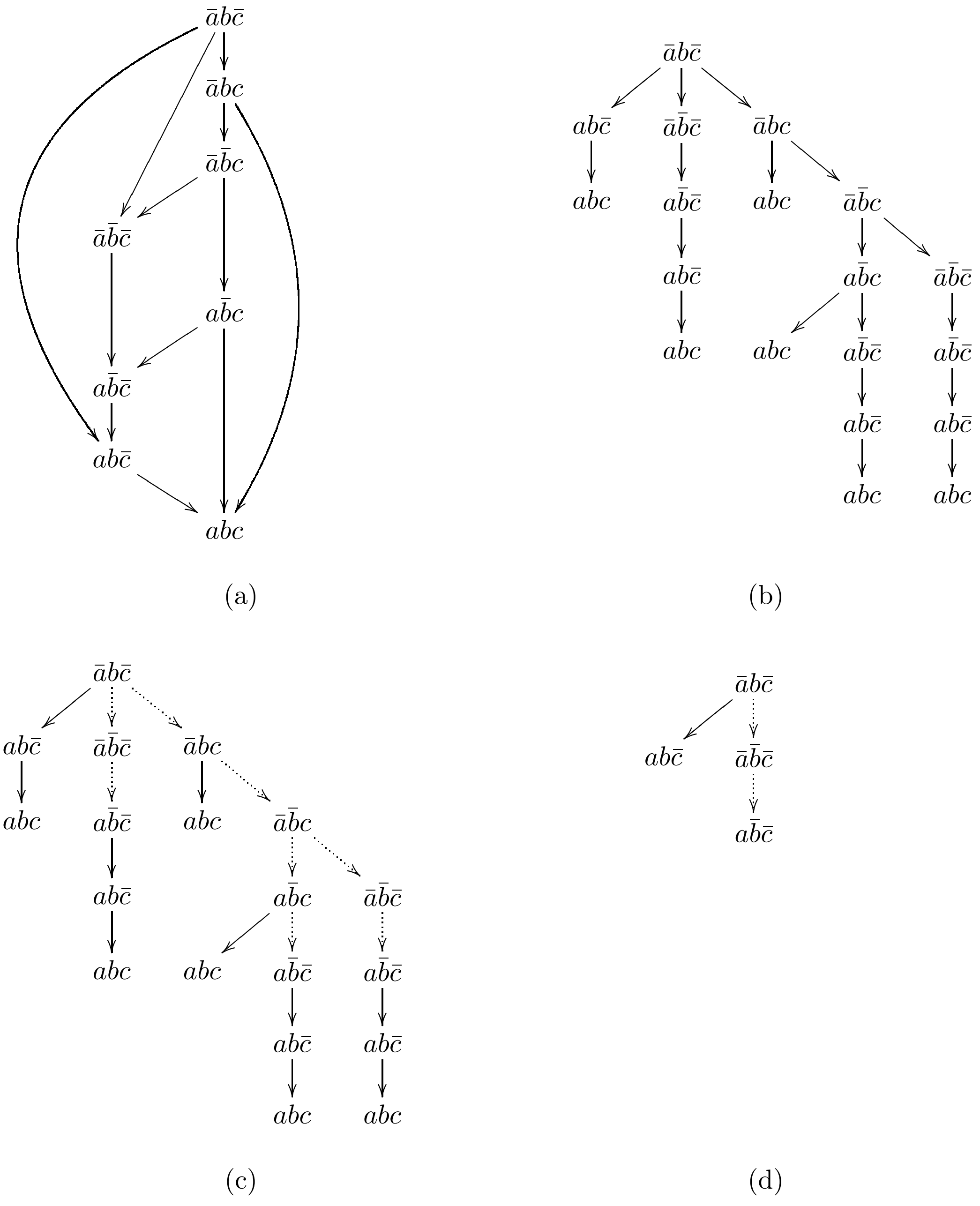

Figure 11: (a) The preference graph induced by a CP-net; (b) the improving search tree $T(\bar{a}b\bar{c})$; (c) paths to outcome $a\bar{b}\bar{c}$; and (d) pruning $T(\bar{a}b\bar{c})$ by suffix fixing.

Suppose, given query $N \models o \succ o'$, there is a path in $T(o')$ to an outcome $o^*$ of the type mentioned in the lemma above. Since the subtree of $T(o')$ rooted at $o^*$ is just $T(o^*)$, we are assured that if any path from $o'$ to $o$ passes through $o^*$, there is a path from $o'$ to $o$ that passes through $o^*$ in which the suffix of $o^*$ is preserved on the subpath from $o^*$ to $o$. From this we conclude:

**Proposition 18** *Any complete search algorithm for the improving search tree remains complete if pruning using the suffix rule is used.*

The suffix fixing rule effectively prunes the search tree under a node $o^*$ as described above to contain only paths that retain the suffix values of the target $o$. Though backtracking over choices that lead to $o^*$ may be required, as may those choices at $o^*$ that preserve suffixes, consideration of the full search tree under $o^*$ will not be required (if there is a nontrivial suffix match). For example, the improving search tree $T(\bar{a}b\bar{c})$ for the query $N \models a\bar{b}\bar{c} \succ \bar{a}b\bar{c}$ discussed above, pruned using the suffix fixing rule and the variable ordering $A, B, C$, appears in Figure 11(d). As we can see, this pruning can dramatically reduce the size of the effective search tree.

## 5.2 Least-Variable Flipping

An extension of the suffix fixing rule is the *least-variable flipping rule*, defined as follows. Suppose we have a CP-net $N$, and a query $N \models o \succ o'$. Let $o^*$ be an outcome, and for any variable $X_j$, let $\mathbf{u}$ denote the instantiation of $\mathbf{U} = Pa(X_j)$ in $o^*$. We say $X_j$ is a *least-improvable variable in* $o^*$ if there is some value $x \in Dom(X_j)$ such that $x \succ^j_{\mathbf{u}} o^*[X_j]$, and no descendent of $X_j$ in $N$ has this property. Intuitively, a least-improvable variable is one that is the lowest in $N$ that can be flipped to produce an outcome that is preferred to $o^*$. Naturally, there may be several such variables. We say $X_j$ is least improvable *with respect to the target* $o$ iff $X_j$ is the least improvable variable among those not part of a suffix match with $o$. In other words, if some suffix of $o$ and $o^*$ matches, we apply the definition of least-improvable variable, restricting attention to those variables that are not part of the matching suffix. The least-variable flipping rule requires that the only neighbors of a node $o^*$ that can be expanded in the search for an improving sequence with target $o$ are those in which some least improvable variable with respect to $o$ is improved.

The following lemma ensures that for binary-valued, directed-path singly connected CP-nets, pruning an improving search tree using the least-variable rule does not affect the completeness of any search procedure.

**Lemma 19** *Let $N$ be a binary-valued, directed-path singly connected CP-net, and $o$ and $o^*$ be outcomes whose $r$th suffix matches (for any topologically consistent ordering $X_1, \cdots, X_n$ of the variables in $N$). Let $\{o_1, \ldots, o_m\}$ $(m \leq n)$ be the set of all outcomes reachable from $o^*$ via some least-variable flip that does not affect the matched suffix. If there is path in $T(o^*)$ from root $o^*$ to $o$, then there is a path from some $o_s$ $(s \leq m)$ to $o$.*

**Proof:** Without loss of generality and based on our earlier observations, we can assume no suffix match between $o$ and $o^*$ (if so, restricting attention to the set of non-suffix variables does not affect our argument).

Now assume, contradicting the statement of the theorem, that none of the least-variable flips $o_s$ of $o^*$ have a path to $o$, but there does exist a path from $o^*$ to $o$. This implies that none of these least-variable flips involve leaf variables in the network $N$. Otherwise, we could flip the value of such a leaf variable at $o^*$ without *any* effect on our ability to flip other variables, and thus be able to construct a path from $o^*$ to $o$ that passes through any of the $o_s$ that flip leaves.

Now, consider a leaf variable $X_i$ in $N$. Since we are dealing with binary variables, on any irreducible flipping sequence from $o^*$ to $o$ the value of $X_i$ should be flipped exactly once. However, the current assignment $o^*[Pa(X_i)]$ does not allow us to perform this flip (see observation above). Thus, we must achieve another assignment to $Pa(X_i)$ before we can flip the value of $X_i$, making it a part of a suffix match.

Let $N_{X_i}$ be the subnetwork of $N$ induced by $X_i$ and all ancestors of $X_i$ in $N$. Because $N$ is directed-path singly connected, $N_{X_i}$ forms a tree, directed toward its root $X_i$. Now we reduce $N_{X_i}$ further by removing all subtrees $N_{X_j}$ of $N_{X_i}$, such that no variable in $N_{X_j}$ can be flipped in $o^*$. Note that this step cause no loss of generality, since (due to acyclicity of $N$) the assignment $o^*$ to the variables in $N_{X_j}$ cannot be flipped first in any improving flipping sequence beginning with $o^*$. Likewise, let $X_{i_1}, \ldots, X_{i_{m'}}$, $m' \leq m$, be the variables corresponding to the least-variable flips of $o^*$ in $N_{X_i}$, and, for $1 \leq k \leq m'$, let $\rho_{i_k}$ be the (single) directed path from $X_{i_k}$ to $X_i$.

Let $\Upsilon = \bigcup_{j=1}^{m'} \left( N_{X_{i_j}} \setminus \{X_{i_j}\} \right)$. Because $N_{X_i}$ is a tree, for $1 \leq j \leq m'$, the variable $X_{i_j}$ *separates* the variables in $N_{X_{i_j}} \setminus \{X_{i_j}\}$ and the paths $\rho_{i_1}, \ldots, \rho_{i_{m'}}$. Therefore:

(i) No set of flips of the variables in $\Upsilon$ will enable us to flip the values of the variables on $\rho_{i_1}, \ldots, \rho_{i_{m'}}$; in particular a flip of $Pa(X_i)$ cannot be enabled. Thus, to enable a flip of $X_i$, eventually we will have to flip the value of at least one variable from $X_{i_1}, \ldots, X_{i_{m'}}$.

(ii) No value flip of $X_{i_1}, \ldots, X_{i_{m'}}$ will affect (neither positively nor negatively) our ability to flip the values of the variables in $\Upsilon$. Thus, if there is an improving flipping sequence from $o^*$ to $o$, then at least one such sequence starts with a value flip of a variable from $X_{i_1}, \ldots, X_{i_{m'}}$.

This last observation contradicts the assumption that there is a flipping sequence from $o^*$ to $o$ that does not pass through at least one $o_s$. □

The least-variable flipping rule does not distinguish the flips of different candidate least improvable variables; it simply restricts flips to such variables. In general, not all least-variable flips are suitable—some may lead to dead-ends, requiring backtracking (a point illustrated in Section 4.4.3). However, when we backtrack, we need only consider other least-variable flips, not all flips, thus significantly reducing the size of the search tree and the expected amount of backtracking. We observe that the TreeDT algorithm for binary-valued, tree-structured CP-nets essentially implements the least-variable flipping rule, which is, therefore, a complete and backtrack-free search procedure for binary-valued, tree-structured networks.

Examples 7 and 8 below show that Lemma 19 presents probably the widest class of CP-nets for which the least-variable flipping rule is complete: Example 7 shows that the least-variable flipping rule does not preserve completeness in binary-valued, max-2-connected

CP-nets, while Example 8 shows the same for multi-valued, directed-path singly connected CP-nets. Note that the CP-net in Example 8 forms a chain. Therefore, binary-valued, tree-structured CP-nets are probably the widest class of CP-nets for which the least-variable flipping rule is both complete and backtrack-free.

**Example 7** Consider the binary-valued CP-net in Figure 12, where Figure 12(a) depicts the graph of the CP-net, and Figure 12(b) shows the corresponding CPTs. Given query $N \models abcde \succ \bar{a}\bar{b}\bar{c}\bar{d}\bar{e}$, we work in the improving tree rooted at $\bar{a}\bar{b}\bar{c}\bar{d}\bar{e}$. The only least improvable variable that can be flipped at the root outcome is $B$ (while $A$ and $E$ can be flipped, they are not least improvable). Unfortunately, flipping $B$ to value $b$ leads to outcome $\bar{a}b\bar{c}\bar{d}\bar{e}$, from which the target $abcde$ is unreachable:

(i) In order to reach the target value $d$ for the variable $D$, first we should achieve the assignment $\bar{b} \wedge c$ to the variables $B$ and $C$;

(ii) Achieving the assignment $\bar{b} \wedge c$ after the flip in question from $\bar{b}$ to $b$ (given $\bar{a} \wedge \bar{e}$) is possible only by restoring the value $\bar{b}$ for $B$, which requires $a \wedge e$; and

(iii) Flipping the value of $B$ from $b$ to $\bar{b}$ given $a \wedge e$ will lead us to a situation in which we can no longer flip $B$, preventing us from achieving the target value $b$.

Thus the least-variable flipping rule will not allow the discovery of an improving flipping sequence.

On the other hand, there is a flipping sequence from $\bar{a}\bar{b}\bar{c}\bar{d}\bar{e}$ to $abcde$ that proves the query:

$$\bar{a}\bar{b}\bar{c}\bar{d}\bar{e} \rightarrow a\bar{b}\bar{c}\bar{d}\bar{e} \rightarrow a\bar{b}c\bar{d}\bar{e} \rightarrow a\bar{b}cd\bar{e} \rightarrow abcd\bar{e} \rightarrow abcde$$

This sequence requires that $A$ be flipped at the root, despite the fact that it is not a least-improvable variable. □

**Example 8** Consider a chain CP-net with three variables $A$, $B$, and $C$, such that $A$ is the parent of $B$, and $B$ is the parent of $C$. Suppose $A$ has domain $\{a, \overline{a}\}$, $B$ has domain $\{b_1, b_2, b_3\}$, and $C$ has domain $\{c, \overline{c}\}$, with the following conditional preferences:

$$a \succ \overline{a};$$
$$a : b_3 \succ b_2 \succ b_1;$$
$$\overline{a} : b_3 \succ b_1 \succ b_2;$$
$$b_1 : c \succ \overline{c} \quad b_2 : \overline{c} \succ c; \quad b_3 : c \succ \overline{c}$$

Consider the query $N \models ab_3\overline{c} \succ \overline{a}b_1c$. The value $c$ cannot be improved in the context of $b_1$, but $b_1$ can be improved to $b_3$ in the context of $\overline{a}$; in fact this is the only least-improvable variable flip for outcome $\overline{a}b_1c$. Unfortunately, this flip leads to outcome $\overline{a}b_3c$ from which no path to the target $ab_3\overline{c}$ exists. In contrast, flipping the non-least-improving variable $A$ to $a$ first allows the discovery of a successful improving path: after flipping $\overline{a}$ to $a$, we change $b_1$ to $b_2$, $c$ to $\overline{c}$, and finally $b_2$ to $b_3$. □

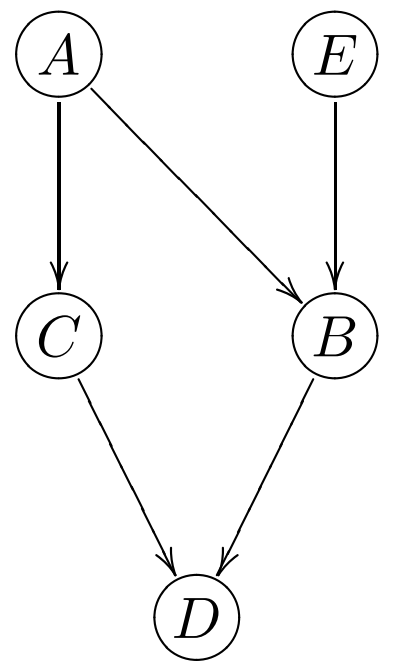

(a)

| Variable | CPT |
|---|---|
| $A$ | $a \succ \bar{a}$ |
| $B$ | $a \wedge e \ : \ \bar{b} \succ b$<br>$\bar{a} \vee \bar{e} \ : \ b \succ \bar{b}$ |
| $C$ | $a \ : \ c \succ \bar{c}$<br>$\bar{a} \ : \ \bar{c} \succ c$ |
| $D$ | $c \wedge \bar{b} \ : \ d \succ \bar{d}$<br>$\bar{c} \vee b \ : \ \bar{d} \succ d$ |
| $E$ | $e \succ \bar{e}$ |

(b)

Figure 12: A multiply connected CP-net for which the least-variable flipping rule causes incompleteness.

Though for multiply-connected networks, and networks with multi-valued variables, the least-variable flipping rule is not complete, we believe that it can provide useful heuristic guidance in these cases. The *least-variable-first heuristic* is a heuristic for ordering children in an improving search tree—it requires that when expanding a node $o^*$ (with respect to some target $o$), the children corresponding to least-improving variables be explored first. This will typically reduce the number of nodes expanded in the tree, because the least-variable-first heuristic can be viewed as embodying a form of *least commitment*. Flipping the values of a least-improving variable can be seen as leaving maximum flexibility in flipping the values of other variables. An upstream variable limits the possible flipping sequences more drastically than a downstream variable since altering a variable does not limit the ability to flip the values of its non-descendants.

For multivalued networks, the least commitment strategy can be used to extend the least-variable-first heuristic using the *least improving* rule: alternative improving flips of the same least-improvable variable are considered from the least improving flip to the most improving flip (i.e., first the flip that leads to the *least preferred* improving value is adopted). This allows greater flexibility in the movement of "downstream" variables. While one can always further improve the value of the variable in question from its less preferred value to a more preferred value (provided that parent values are maintained), "skipping" values may prevent us from setting descendants to their desired values. In fact, this was illustrated in Example 8, where, after the crucial flip of variable $A$, using the *least improving* rule as a heuristic leads directly to the target outcome.

### 5.3 Forward Pruning

In a search procedure for an improving flipping sequence, no matter whether this procedure adopts the above heuristics or not, one can use a general *forward pruning* technique. This technique has a number of desirable properties:

(a) it often quickly shows that no flipping sequence is possible;

(b) it prunes the domains of the variables to reduce the flipping search space;

(c) it doesn't compromise soundness or completeness; and

(d) it is relatively cheap: its time complexity is $O(nrd^2)$, where $n$ is the number of variables, $r$ is the maximum number of conditional preference rules for each variable, and $d$ is the size of the largest variable domain.

The general idea is to sweep forward through the network, pruning any values of a variable that cannot appear in any improving flipping sequence for a given query. Intuitively, we consider the set of flips possible, ignoring interdependence of the parents and the number of times the parents can change their values.

We consider the variables in an order consistent with the network topology (so that parents of a node are considered before the node). For each variable $X_j$ with parents $\mathbf{U}$, we build a *domain transition graph* with nodes corresponding to the possible values $x_i \in Dom(X_j)$. For each conditional preference relation $\succ^j_{\mathbf{u}}$ over $Dom(X_j)$ of the form:

$$x^{\mathbf{u}}_1 \succ x^{\mathbf{u}}_2 \succ \cdots \succ x^{\mathbf{u}}_d$$

such that $\mathbf{u}$ contains only unpruned values of the parents $\mathbf{U}$ of $X_j$, we include directed arcs between the successive values in the ordering $\succ^j_{\mathbf{u}}$ (i.e., an arc from $x_i$ to $x_{i-1}$, for each $1 < i \leq d$).

When answering query $N \models o \succ o'$, we can prune any value of $X_j$ that is not on a directed path from $o'[X_j]$ to $o[X_j]$ in the domain transition graph for $X_j$. This can be implemented by running the well-known Dijkstra's algorithm (Cormen, Lierson, & Rivest, 1990) twice: once to find the nodes reachable from $o'[X_j]$ and again to find the nodes that can reach $o[X_j]$. These sets of nodes can be intersected to find the possible values for $X_j$ along any path from $o'[X_j]$ to $o[X_j]$ in $T(o')$ (i.e., along any improving sequence from $o'[X_j]$ to $o[X_j]$ with respect to $N$). If the intersection is empty, the dominance query fails: there is no legal flipping sequence from $o'[X_j]$ to $o[X_j]$. This often results in quick failure for straightforward queries, so that we only carry out the search for non-obvious cases.

**Example 9** Consider any CP-net $N$ in which $A$ and $B$ are both binary root variables, and the preferences over the values of $A$ and $B$ are $a \succ \bar{a}$ and $b \succ \bar{b}$. Given a query $N \models a\bar{b} \ldots \succ \bar{a}b \ldots$, first we consider $A$. The domain transition graph for $A$ consists of an arc $\bar{a} \rightarrow a$, thus no values of $A$ are pruned. If the example were changed slightly so that $A$ had a third value $\overline{\overline{a}}$, where $a \succ \overline{a} \succ \overline{\overline{a}}$, then this third value could be pruned from $A$, thus simplifying the tables for all the children of $A$.

We then consider $B$, whose domain transition graph consists of a single arc $\bar{b} \rightarrow b$. Since the value of $B$ in the (purported) more preferred outcome of the query ($\bar{b}$) is not reachable in the domain transition graph of $B$ from the value of $B$ in the (purported) less preferred outcome of the query ($b$), the query fails quickly without looking at the other variables. □

## 6. Incompletely Specified Preferences and Indifference

In many practical applications, one expects to see reluctance on behalf of the user to provide complete CPTs or to totally order the values of each variable in every possible context. Thus, it is natural to ask how our results and techniques can be applied in these cases.

It turns out that all the results presented in this paper, except for the linear time procedure for ordering queries, can be easily extended to work both on CP-nets with partially specified CPTs, and on satisfiable CP-nets that capture statements of preferential indifference. For instance, all the results presented in Section 4.4 with respect to *dominance queries* remain applicable "as is," because almost all of these results were shown by Brafman and Domshlak (2003) to be valid in more general setting of classical planning. The only case that was not analyzed by Domshlak and Brafman is the case of tree CP-nets. However, the correctness of the TreeDT procedure for "extended" CP-nets can be easily verified, and its complexity remains quadratic. This stems from the fact that Lemma 10 is valid in the more general planning setting (Brafman & Domshlak, 2003). The only point that requires clarification is the difference between partial specification and indifference with respect to flipping sequences: Given a variable $X$ with parents $\mathbf{U}$, and two values $x_1, x_2 \in Dom(X)$, if $x_1$ and $x_2$ are equally preferred given $\mathbf{u} \in Asst(\mathbf{U})$, then given $\mathbf{u}$ we can flip the value of $X$ from $x_1$ to $x_2$, and vice versa. Alternatively, if $x_1$ and $x_2$ are incomparable given $\mathbf{u} \in Asst(\mathbf{U})$, then given $\mathbf{u}$ we cannot flip the value of $X$ from $x_1$ to $x_2$, nor from $x_2$ to $x_1$.

The complexity of dominance testing in the context of indifference and incompletely specified CPTs remains an open problem. However, Theorem 20 below shows that the flipping sequence search over multi-valued CP-nets with partially specified preferences is not in NP, even if the CP-net forms a *chain* and the variables are three-valued only.

**Theorem 20** *Flipping sequence search over multi-valued CP-nets with partially specified preferences is not in* NP.

**Proof:** For the proof see Appendix A.

Now, consider *outcome optimization* queries. When the CPTs are allowed to be partially specified, or the statements of indifference are allowed, the CP-net may induce more than one nondominated outcome. For instance, consider again Example 1 (My Dinner I). If the preference for the type of wine given fish soup is not specified, or the decision maker considers both red and white wine to go equally well with fish soup, then this CP-net induces two nondominated outcomes: fish soup with white wine, and fish soup with red wine.

The forward sweep procedure for the outcome optimization queries presented in Section 3 can be easily extended for the cases of partial specification and/or indifference by adding branching on each variable $X$ for which the (already generated) assignment on $Pa(X)$ induces more than one nondominated value from $Dom(X)$. The complexity of the resulting algorithm is $O(n\alpha)$, where $n$ is the number of variables in the CP-net, and $\alpha$ is the number of nondominated outcomes induced by the CP-net. Of course, $\alpha$ can be exponential in the size of the CP-net.[11] However, our adapted forward sweep procedure has the *anytime* property—the solutions are generated iteratively, and the time to add a new nondominated

11. The tight upper bound of $\alpha = 2^n$ can be shown for any CP-net by simply leaving all CPTs unspecified.

outcome to the current set of generated solutions is $O(n)$. Therefore, the complexity of generating $k$ nondominated solutions is linear in $k$.

Finally, an important query that can be answered efficiently in standard CP-nets is the *outcome ordering* query (see Section 4.1). Although the basic Corollary 4 remains valid in the case of CP-nets with partial specifications or statements of indifference, this is not the case for Theorem 5. Therefore, the computational complexity of ordering queries for extended CP-nets remains an open question, and we conjecture that this problem is hard.

## 7. Concluding Remarks

In this paper we introduced CP-nets, a new graphical model for representing qualitative preference orderings which reflects conditional dependence and independence of preference statements under a *ceteris paribus* semantics. This formal framework offers a compact and arguably natural representation of preference information, and allows us to efficiently answer some of the principal forms of preference queries.

We described several types of queries and algorithms for answering them with respect to a specific CP-net. In particular, outcome optimization and outcome ordering queries were shown to be solvable in time linear in the number of variables in the network. For the dominance queries, however, the situation is more complicated. First, we demonstrated the equivalence of answering dominance queries with the task of determining existence of an improving (or worsening) sequence of variable value flips with respect to the given CP-net. Then, we reduced the latter task to a special subclass of classical planning problems. These insights allowed us to show that, in general, answering dominance queries is NP-hard, but that polynomial algorithms exist for tree and polytree-structured, binary-valued CP-nets.

In addition, we presented several techniques that one can use in a generic search procedure for an improving flipping sequence. Some of these techniques were shown to have no impact on soundness or completeness of the search for any CP-net, while other techniques have this property only for binary-valued CP-nets. However, we argued that latter techniques can be modified into general purpose heuristics that are likely to reduce significantly the size of the expanded search tree.

Finally, we analyzed the applicability of our results for CP-nets that allow partially specified preferences and/or capture statements of indifference.

### 7.1 Applications

Our goal in developing the CP-nets formalism is to facilitate the development of applications. One such application—preference-driven, adaptive multimedia document presentation—was recently developed at Ben-Gurion University (Domshlak et al., 2001; Gudes et al., 2002). We described the central components of this system in Section 3.2. Another application in which both conceptual and computational properties of CP-nets seem to be useful is a distributed meeting scheduler. A basic prototype for such a system has been implemented at Ben-Gurion University (Brafman & Domshlak, 2001). We are currently extending this system, while working on the related theoretical issue of multi-agent preference-based optimization.

Another potential application for qualitative preferences in general, and thus for CP-nets in particular, is sorting a product database according to user-specified preferences. This problem is highly relevant in the context of electronic commerce. Several rather conceptually simplistic, though quite interesting, commercial applications that rely on unconditional preference statements are available on the World Wide Web; examples include Active Sales Assistant™ (see `www.activebuyersguide.com`) and Personalogic™ (see `www.personalogic.com`). The general idea is to assist a user in selecting a specific product from a product database according to her preferences. Here, it is very important to use compact and natural representations for preference information. CP-nets extend current models (which typically don't allow conditional preference statements), yet offer efficiency in ordering a given set of alternatives. Another important aspect of this problem is that the given database precisely defines the products (represented as vectors of attribute values) available, and preference information is only required to the extent that it narrows the choice of product to a sufficiently small selection of products from this database. Both the graphical properties of CP-nets underlying the efficiency of the ordering queries, and the various dominance testing strategies, can be exploited in this context to find a subset of products that are not dominated by any other product in the database, given the (conditional) preference information extracted from the user. Here, an interactive and dynamic approach appears to be most promising, where the user is prompted for additional preference statements until the ordering of the products in the database is sufficiently constrained by the preference information to offer a reasonably small selection of products.

Another growing application area for CP-nets is automated constraint-based product configuration (Sabin & Weigel, 1998). The task is to assemble a number of components that compose a product such that given compatibility constraints are satisfied. A simple example of this is the assembly of components for a computer system where, for instance, the type of system bus constrains the choice of video and sound cards. While there has been a wide and growing body of research in modeling configuration problems and efficient problem solving methods, there is still a need for more work on modeling and learning user preferences, and using these to achieve configurations that are not only feasible, but also satisfactory from the user point of view. These issues are emphasized in many papers on configuration (Freuder & O'Sullivan, 2001; Haag, 1998; Junker, 1997, 2001; Soininen & Niemelä, 1998), especially when high-level configurators for specific, real-life domains are discussed (Haag, 1998). The importance of incorporating user preferences into the configuration problem stems from the fact that many such problems are weakly constrained and have numerous feasible solutions (D'Ambrosio & Birmingham, 1995). The value of these solutions, from the subjective point of view of a particular user, may vary significantly.

CP-nets can be used to represent user preferences which will be used together with compatibility constraints to search for most preferred, feasible configurations. In contrast to the database sorting application above, here the set of possible vectors of feature values (i.e., configurations) is not explicitly given, but implicitly specified by the compatibility constraints. A CP-net based search algorithm by Boutilier et al. (1997) was specifically designed to address this problem. For the description of this algorithm, as well as for analysis of its computational properties we refer the reader to Boutilier et al. (1997), Brafman and Domshlak (2001).

### 7.2 Related Work

A number of lines of research are related to CP-nets. In addition to the conceptual work in philosophy and philosophical logic described in Section 2, in AI, Doyle and Wellman (1991, 1994) explored *ceteris paribus* assertions and their logical properties. However, their work did not exploit the notions of preferential independence, and in particular did not considered graphical representations of preference statements. To the best of our knowledge, there are no computational results known for this formalism. Therefore, it is not clear whether useful queries can be answered efficiently in this framework.

On the surface, CP-nets are reminiscent of Bayesian networks (Pearl, 1988), which are also graphical structures capturing conditional independence assertions. Indeed, Bayesian networks and their utilization of probabilistic independence provide important motivation to our work, but the two structures differ considerably in their properties and the type of information they present.

Motivated by the same considerations driving our work, Bacchus and Grove (1995) and La Mura and Shoham (1999) study different notions of independence and their associated graphical representations. Both representations allow for quantitative assessments, unlike CP-nets (at least in their current form) and differ from CP-nets in the precise nature of the independence concept studied. In particular, Bacchus and Grove concentrate on the notion of *conditional additive independence.* Additive independence is a very strong property, requiring that the utility of an outcome be the sum of the "utilities" of the different variable values of the outcome. Conditional additive independence is a weaker requirement, and thus more promising in practice. Bacchus and Grove show that the conditional additive independence properties of a domain can be captured by an undirected graph where for set of nodes $\mathbf{A}, \mathbf{B}, \mathbf{C}$, $\mathbf{A}$ is independent of $\mathbf{B}$ given $\mathbf{C}$ if $\mathbf{C}$ separates between the nodes in $\mathbf{A}$ and $\mathbf{B}$. La Mura and Shoham (1999) define the concept of *u-independence* using a *ceteris paribus* comparison operator over utilities. This operator measures the "intensity" of preference for specific values of certain variables given some fixed value for the other variables and with respect to a fixed reference point. They also define an undirected graphical structure, *expected utility networks*, in which u-independence is represented using the notion of node separation.

Finally, recent work by Benferhat, Dubois and Prade (2001) provides a preliminary investigation of the potential of possibilistic logic in qualitative decision analysis, and more specifically in qualitative preference representation. The possibilistic approach takes utilities into account, as well as probabilities, but provides a qualitative approach to decision problems by replacing numeric preferences and probabilities with linear orderings. For discussion of this approach to utilities and preference representation, see Benferhat et al. (2001), Dubois, Godo, Prade, and Zapico (1998b), Dubois, Berre, Prade, and Sabbadin (1998a), Dubois, Prade, and Sabbadin (1997), and for related qualitative models, see Boutilier (1994), Tan and Pearl (1994).

### 7.3 Future Work

We see a number of potential extensions to the work described in this paper. In particular, our work leaves a number of interesting open theoretical questions. First, the worst case complexity of dominance testing with respect to acyclic, binary-valued CP-nets needs to be established—it is still an open question whether this problem is in NP. Second, the complexity of dominance queries with respect to multi-valued CP-nets remains an open problem. Third, it is not clear how hard ordering queries are for CP-nets that capture partial preference specification and/or the statements of preferential indifference. Finally, there is a need to understand the expressive power of CP-nets better, specifically: what sort of partial orderings are and are not representable by CP-nets; and among orderings representable by CP-nets, which ones are can only be represented by a cyclic network.

Cyclic networks present another important challenge. Such networks can arise in applications where a natural notion of neighborhood is defined on the set of variables such that preferences for one variable value depend on the value of its neighboring variables. In such cases, the user may find it more natural to provide a cyclic preference structure and we must be able to determine whether the specified network is satisfiable. In addition, inference methods for such networks need to be developed or, alternatively, methods for reducing cyclic networks to acyclic networks without significant blow-up in size. A number of recent papers have started to deal with this question, with some interesting insights and results (Domshlak, 2002a; Domshlak & Brafman, 2002a; Brafman & Dimopoulos, 2003; Domshlak, Rossi, Venable, & Walsh, 2003).

We are working on various extensions of the framework presented here. These extensions include the use of conditional preference statements that contain a small amount of quantitative information. Existing applications (such as online interactive consumer guides) suggest that a limited amount of such quantitative preference information might be relatively easy to extract from the user in a natural way, and is very useful for inducing stronger preference orderings. Preliminary work on this topic has been undertaken (Boutilier, Bacchus, & Brafman, 2001).

Another interesting extension of CP-nets are TCP-nets (Brafman & Domshlak, 2002). TCP-nets add importance relations and conditional relative importance statements to the the conditional *ceteris paribus* statements supported by CP-nets. Conditional importance statements have the form "if $A = a$ then optimizing $B$ is more important to me than optimizing $C$". For example, "If I will be flying at night, then having a better seat is more important than getting a more preferred carrier."

Finally, preference-based optimization presents an interesting tradeoff between the amount of user interaction required for extracting preference information and the amount of computation needed to determine the most preferred feature vectors. By asking very specific questions about particular, potentially complex preferences, finding most preferred feature vectors can become much easier. On the other hand, asking too many questions, especially those not really necessary for establishing relevant preferences, will annoy the user and make the system less usable. Thus, finding good tradeoffs between the amount of user-interaction and computation time for answering queries—such as finding most preferred items from a database of optimal configurations—seems to be a promising direction for future research. This is related to the motivation underlying goal programming (Dyer, 1972; Ignizio, 1982). The structure of CP-nets can be exploited in determining preference elicitation strategies. This has been explored in the context of CP-nets with quantitative information (Boutilier et al., 2001); it remains to be seen how to use such techniques in a purely qualitative setting.

## Acknowledgments

Some parts of this paper appeared in C. Boutilier, R. Brafman, H. Hoos, and D. Poole, "Reasoning with Conditional Ceteris Paribus Preference Statements," *Proceedings of the Fifteenth Conference on Uncertainty in Artificial Intelligence*, pp.71–80, Stockholm, 1999; and in C. Domshlak and R. Brafman, "CP-nets—Reasoning and Consistency Testing," *Proceedings of the Eighth International Conference on Principles of Knowledge Representation and Reasoning*, pp.121–132, Toulouse, 2002. Thanks to Chris Geib for his contributions to earlier, related models of CP-nets, and to the anonymous referees for their many useful suggestions. Boutilier, Hoos and Poole were supported by the Natural Sciences and Engineering Research Council, and the Institute for Robotics and Intelligent Systems. Brafman acknowledges the support of the Paul Ivanier Center for Robotics Research and Production Management.

## Appendix A.

**Theorem 13** $\Theta(n^2)$ *is a lower bound for the flipping-sequence search over binary-valued, tree-structured CP-nets.*

**Proof:** The proof is by example of a dominance testing query over a binary-valued, tree CP-net for which the size of a minimal flipping sequence is $\Theta(n^2)$.

Consider the following CP-net $\mathcal{N}$ defined over binary variables $\{X_1, \ldots, X_n\}$, where $n = 2k + 1$ for some $k \in \mathbb{N}$. The domain of each variable $X_i$ is $\mathcal{D}(X_i) = \{x_i, \bar{x}_i\}$. For $1 \leq i \leq n$, $Pa(X_i) = \{X_{i-1}\}$, thus $\mathcal{N}$ forms a directed chain. The CPTs capturing the preferences over the values of $\{X_1, \ldots, X_n\}$ are shown in Figure 13(a).

Now consider the dominance testing query $[\![N \models o \succ o']\!]$, where, for $1 \leq i \leq n$:

$$\begin{aligned} o[X_i] &= x_i \\ o'[X_i] &= \begin{cases} x_i, & i = 2m, \\ \bar{x}_i, & i = 2m+1 \end{cases} \qquad m \in \mathbb{N} \end{aligned}$$

The length of the minimal (and actually the only) flipping sequence from $o'$ to $o$ is $k^2+2k+1$, proving the required lower bound. The $j$th flip on this sequence changes the value of the variable $X_{k+1-\alpha+\beta}$, where:

$$\begin{aligned} \alpha &= \lfloor j/(k+1) \rfloor \\ \beta &= (j-1) \bmod (k+1) \end{aligned}$$

Informally, the first $k+1$ flips change the values of $X_{k+1}, \ldots, X_n$ (in this order), next $k+1$ flips change the values of $X_k, \ldots, X_{n-1}$, then $X_{k-1}, \ldots, X_{n-2}$, etc. Finally, the last $k+1$ flips change the values of $X_1, \ldots, X_{k+1}$. There are $k+1$ such sets, of $k+1$ flips each, the length of the above flipping sequence is thus $k^2 + 2k + 1$.

Figure 13(b-c) illustrates $o$ and $o'$, and the corresponding flipping sequence for $k = 2$ (i.e., $n = 5$). In Figure 13(c), each variable $X_i$ is annotated with its value flips along the flipping sequence: Arrows between the values stand for the value flips, and the sequential numbers of the flips along the flipping sequence appear as the arrow labels.

| Variable | CPT |
|---|---|
| $i = 1$ | $x_i \succ x_i'$ |
| $1 < i \leq k+1$ | $x_{i-1} : x_i \succ \bar{x}_i$<br>$\bar{x}_{i-1} : \bar{x}_i \succ x_i$ |
| $k+1 < i \leq n$ | $x_{i-1} : \bar{x}_i \succ x_i$<br>$\bar{x}_{i-1} : x_i \succ \bar{x}_i$ |

(a)

$$o = x_1 x_2 x_3 x_4 x_5$$
$$o' = \bar{x}_1 x_2 \bar{x}_3 x_4 \bar{x}_5$$

(b)

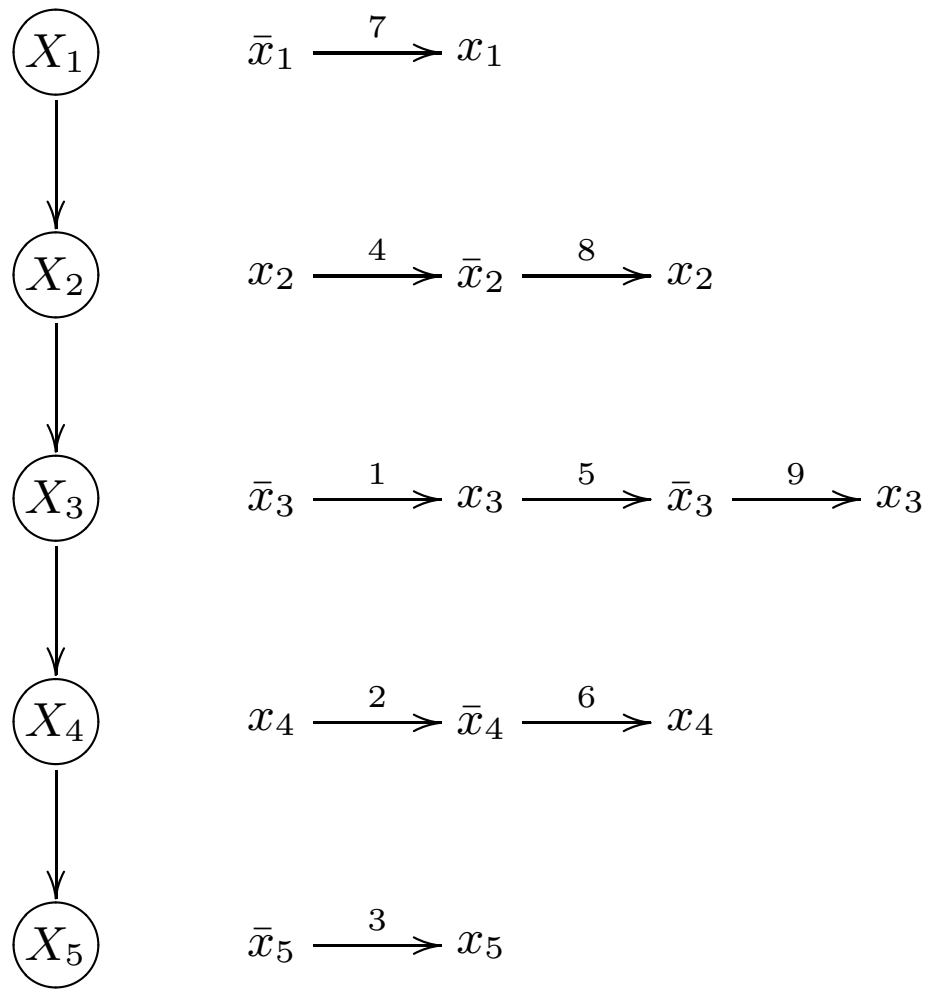

(c)

Figure 13: Illustration for the proof of Theorem 13: (a) CPTs for $n$ variables; (b) $o$ and $o'$ for $k = 2$; and (c) flipping sequence for $k = 2$.

Now we prove that the size of the minimal flipping sequence for this example (with $n = 2k+1$) is $k^2 + 2k + 1$. We divide the proof into two steps, and show that on a minimal flipping sequence from $o'$ to $o$:

1. For $1 \leq i \leq k+1$, every variable $X_{k+i}$ (last $k+1$ variables) must change its value at least $k+2-i$ times, and
2. For $1 \leq j \leq k$, every variable $X_j$ (first $k$ variables) must change its value at least $j$ times.

(1) The proof of the first claim is by induction on $i$. For $i = k+1$, the variable $X_{2k+1}$ (i.e., $X_n$) must change its value at least once since $o[X_{2k+1}] \neq o'[X_{2k+1}]$.

Now, we assume the induction hypothesis that, for $l \leq i \leq k+1$, every variable $X_{k+i}$ must change its value at least $k+2-i$ times, and prove the claim for $X_{k+l-1}$. Recall that $X_{k+l-1}$ is the only parent of $X_{k+l}$, thus every pair of successive flips of $X_{k+l}$ must require different values of $X_{k+l-1}$.

Assume that $k+l$ is even. Then, the first flip of $X_{k+l}$ is from $x_{k+l}$ to $\bar{x}_{k+l}$, requiring $X_{k+l-1}$ to take the value $x_{k+l-1}$ while $o'[X_{k+l-1}] = \bar{x}_{k+l-1}$. Therefore, the first flip of $X_{k+l}$ may be performed only after the first flip of $X_{k+l-1}$, and therefore, before the flip number $k+2-l$ of $X_{k+l}$, the variable $X_{k+l-1}$ will have to change its value at least $k+2-l$ times. Now, since $k+l$ is even, then so is $k+2-l$. Thus, after $k+2-l$ flips, the variable $X_{k+l-1}$ will be assigned the value $o'[X_{k+l-1}] = \bar{x}_{k+l-1}$. However, $o[X_{k+l-1}] = x_{k+l-1}$, thus $X_{k+l-1}$ will have to change its value at least one more time.

The proof for the case of $k+l$ odd is similar: The first flip of $X_{k+l}$ is from $\bar{x}_{k+l}$ to $x_{k+l}$, requiring $X_{k+l-1}$ to take the value $\bar{x}_{k+l-1}$ while $o'[X_{k+l-1}] = x_{k+l-1}$. Therefore, before the flip number $k+2-l$ of $X_{k+l}$, the variable $X_{k+l-1}$ will have to change its value at least $k+2-l$ times. Now, since $k+l$ is odd, then so is $k+2-l$. Thus, after $k+2-l$ flips, the variable $X_{k+l-1}$ will be assigned to the value $\bar{x}_{k+l-1}$. However, $o[X_{k+l-1}] = o'[X_{k+l-1}] = x_{k+l-1}$, thus $X_{k+l-1}$ will have to change its value at least one more time. Hence, we proved that, for $i = l-1$, $X_{k+i}$ has to change its value at least $k+2-i$ $(= k+3-l)$ times.

(2) The proof of the second claim is by induction on $j$. First, we show that $X_k$ must change its value at least $k$ times. From the first claim we have that $X_{k+1}$ changes its value at least $k+1$ times. Therefore, since every pair of successive flips of $X_{k+1}$ requires different values of $X_k$, the variable $X_k$ must change its value at least $k$ times.

Now, we assume the induction hypothesis that, for $l \leq j \leq k$, every variable $X_j$ must change its value at least $j$ times, and prove the claim for $X_{l-1}$. Again, since every pair of successive flips of $X_l$ requires different values of $X_{l-1}$, the variable $X_{l-1}$ must change its value at least $l-1$ times. □

**Theorem 15** *Dominance testing for binary-valued, directed-path singly connected CP-nets is* NP*-complete.*

**Proof:** First we show the membership in NP. Given a dominance query $\Pi = [\![N \models o \succ o']\!]$, let $\mathsf{MinFS}(\Pi)$ denote the size of a minimal (= shortest) improving flipping sequence from $o'$ to $o$. Using the $\mathsf{MaxFlip}$ property of the variables, the following upper bound for $\mathsf{MinFS}(\Pi)$ is straightforward from the Lemma 10:

$$\mathsf{MinFS}(\Pi) \leq \sum_{X_i \in \mathbf{V}} \mathsf{MaxFlip}(X_i) \leq n^2 \tag{2}$$

Thus, if we guess a minimal improving flipping sequence for a given solvable problem, we can verify it in low order polynomial time.

The proof of hardness is by reduction from 3-SAT, i.e., problem of finding a satisfying assignment for a propositional formula in conjunctive normal form in which each conjunct (clause) has at most three literals.

Let $\mathcal{F} = c_1 \wedge \ldots \wedge c_n$ be a 3-SAT propositional formula, and $x_1, \ldots, x_m$ be the variables used in $\mathcal{F}$. Our equivalent problem can be constructed as follows:

- $\mathbf{V} = \{V_1, \bar{V}_1, \ldots, V_m, \bar{V}_m\} \cup \{C_1, \ldots, C_n\}$, where for $1 \leq i \leq 2m+n$, the domain $\mathcal{D}(X_i) = \{f, t\}$ ($f$ and $t$ stand for *false* and *true*, respectively).
- $Pa(V_i) = Pa(\bar{V}_i) = \emptyset$
- $Pa(C_i) = \{V_{i_1}, \bar{V}_{i_1}, V_{i_2}, \bar{V}_{i_2}, V_{i_3}, \bar{V}_{i_3}\}$, where $x_{i_1}, x_{i_2}$, and $x_{i_3}$ are the variables that participate in the $i$th clause of $\mathcal{F}$.
- Outcome $o'$ assigns $f$ to all variables in $\mathbf{V}$.
- Outcome $o$ assigns $t$ to all variables in $\mathbf{V}$.

For any variable $V_i$ or $\bar{V}_i$, $1 \leq i \leq m$, the value $t$ is (unconditionally) preferred to the value $f$. In turn, for $1 \leq i \leq n$, the value $t$ is preferred to the value $f$ for the variable $C_i$ if there exist a $j$, $1 \leq j \leq 3$, such that:

1. $V_{i_j} \neq \bar{V}_{i_j}$
2. if the literal $x_{i_j}$ (and not $\bar{x}_{i_j}$) belongs to the clause $c_i$ then $V_{i_j} = t$, otherwise $\bar{V}_{i_j} = t$

For all other assignments to $Pa(C_i)$, the value $f$ is preferred to the value $t$ for the variable $C_i$.

The constructed inference problem has a directed-path singly connected, binary-valued CP-net in which for $1 \leq i \leq n$, $|Pa(X_i)| \leq 6$. Let us show that an improving flipping sequence from $o'$ to $o$ exists if and only if a satisfying assignment for $\mathcal{F}$ can be found.

$\Leftarrow$ Given a satisfying assignment $\phi$, the improving flipping sequence is as follows: First, for $1 \leq j \leq m$, if $x_j = t$ in $\phi$, then flip the value of the variable $V_j$ from $f$ to $t$. Otherwise, if $x_j = f$ in $\phi$, then flip the value of the variable $\bar{V}_j$ from $f$ to $t$. The actual ordering of these flips is irrelevant since these variables are mutually independent. Then, for $1 \leq i \leq n$, flip the value of $C_i$ from $f$ to $t$. Here also, the ordering of these flips is not significant. Finally, flip the values of all variables $V_j$ and $\bar{V}_j$ that still have the value $f$.

$\Rightarrow$ Let $\mathcal{S}$ be an improving flipping sequence from $o'$ to $o$. If the value of a variable $C_i$ is flipped on $\mathcal{S}$ while variables $V_j, \bar{V}_j \in Pa(C_i)$ had the values $t, f$, respectively, then variable $x_j$ is set to true in $\phi$. Otherwise, if $C_i$ is flipped on $\mathcal{S}$ while variables $V_j, \bar{V}_j \in Pa(C_i)$ had the values $f, t$, respectively, then variable $x_j$ is set to false in $\phi$. Existence of such a $j$ is ensured by the construction of $CPT(C_i)$.

In turn, for $1 \leq j \leq m$, neither $V_j$ nor $\bar{V}_j$ can achieve the value $f$ after achieving the value $t$. Hence, if there is an outcome $o_1 \in \mathcal{S}$ such that $\{V_j = t, \bar{V}_j = f\} \in o_1$, then there will be no outcome $o_2 \in \mathcal{S}$ such that $\{V_j = f, \bar{V}_j = t\} \in o_2$, and vice versa. This shows that the value of each variable $C_i$ is flipped on $\mathcal{S}$ from $f$ to $t$ in a context consistent with $\phi$. Therefore, for each close $c_i$, there is a literal $l_{i_j} \in c_i$, $1 \leq j \leq 3$, such that $l_{i_j} \in \phi$. $\square$

**Theorem 16** *Dominance testing for binary-valued, max-$\delta$-connected CP-nets, where $\delta$ is polynomially bounded in the size of the CP-net, is* NP*-complete.*

**Proof:** We prove this theorem by showing that, for *any* acyclic, binary-valued CP-net $\mathcal{N}$, and for any variable $X_i \in \mathcal{N}$, we have:

$$\mathsf{MaxFlip}(X_i) \leq 1 \; + \sum_{j=i+1}^{n} \rho(X_i, X_j) \tag{3}$$

where $\rho(X_i, X_j)$ denotes the total number of different, not necessary disjoint, paths from $X_i$ to $X_j$. For simplicity of presentation, we assume that the variables $\{X_1, \ldots, X_n\}$ of $\mathcal{N}$ are numbered according to a topological order induced by the graph of $\mathcal{N}$, and rewrite Eq. 3 into:

$$\mathsf{MaxFlip}(X_i) \leq 1 \; + \sum_{j=i+1}^{n} \rho(X_i, X_j) \tag{4}$$

The proof is by induction on $i$. For $i = n$ it is obvious that $\mathsf{MaxFlip}(X_n) \leq 1$, since $X_n$ is a leaf node in $\mathcal{N}$. Now we assume that the lemma holds for any $i > k$, and prove it for $i = k$. Without loss of generality, assume that there exist at least one variable $X_j$ such that $X_k \in Pa(X_j)$. Otherwise, we simply have that $\mathsf{MaxFlip}(X_k) \leq 1$.

Denote by $\mathsf{succ}(X_k)$ the immediate successors of $X_k$ in $\mathcal{N}$, i.e. $X_{i_k} \in \mathsf{succ}(X_k)$ iff $X_k \in Pa(X_{i_k})$. The proof is straightforward:

$$\begin{aligned}\mathsf{MaxFlip}(X_k) &\overset{Lemma\ 9}{\leq} 1 + \sum_{X_{i_k} \in succ(X_k)} \mathsf{MaxFlip}(X_{i_k}) \leq \\ &\overset{I.H.}{\leq} 1 + |\mathsf{succ}(X_k)| + \sum_{X_{i_k} \in succ(X_k)} \sum_{j=i_k+1}^{n} \rho(X_{i_k}, X_j) = \\ &= 1 + \sum_{j=k+1}^{n} \rho(X_k, X_j)\end{aligned}$$

Now, if a binary-valued CP-net is max-$\delta$-connected, and the variables of a given problem $\Pi$ are considered in a topological ordering induced by $\mathcal{N}$, then from Eq. 4 it follows that, for any variable $X_i \in \mathcal{N}$, we have:

$$\mathsf{MaxFlip}(X_i) \leq \delta n + 1 \tag{5}$$

Let $\mathsf{MinFS}(\Pi)$ denote the size of a minimal (= shortest) improving flipping sequence for $\Pi$. From Eq. 5 follows that $\mathsf{MinFS}(\Pi) \leq \delta n^2 + n$. Hence, if $\delta$ is polynomially bounded in the size of $\Pi$, then we can guess and verify a minimal improving flipping sequence for $\Pi$ in polynomial time, and thus this class of dominance testing queries is in NP. □

**Theorem 20** *Flipping sequence search over multi-valued CP-nets with partially specified preferences is harder than* NP.

**Proof:** The proof is by example of a dominance query over a multi-valued, chain CP-net with partially specified preferences, for which the minimal flipping sequence is exponentially long.

Consider the following CP-net $\mathcal{N}$ defined over three-valued variables $\{X_1, \ldots, X_n\}$, where $n = 2k + 1$ for some $k \in \mathbb{N}$. The domain of each variable $X_i$ is $\mathcal{D}(X_i) = \{x_i, \bar{x}_i, \bar{\bar{x}}_i\}$. For $1 \leq i \leq n$, $Pa(X_i) = \{X_{i-1}\}$, thus $\mathcal{N}$ forms a directed chain. The CPTs capturing the preferences over the values of $\{X_1, \ldots, X_n\}$ are shown in Figure 14.

Now consider the dominance testing query $[\![N \models o \succ o']\!]$, where, for $1 \leq i \leq n$, $o[X_i] = x_i$ and $o'[X_i] = \bar{\bar{x}}_i$. Let the sequence $\{a_i\}$ be defined as:

$$a_i = \begin{cases} 2, & i = 1 \\ 2a_{i-1} + 2, & i \geq 2 \end{cases} \tag{6}$$

The length of a minimal flipping sequence from $o'$ to $o$ as above is greater than:

$$\sum_{i=1}^{k} a_i > 2^{\frac{n}{2}}$$

Figure 15 illustrates the corresponding flipping sequence for $k = 2$ (i.e., $n = 5$), where each variable $X_i$ is annotated with its value flips along the flipping sequence: Arrows between the values stand for the value flips, and the sequential numbers of the flips along the flipping sequence appear as the arrow labels.

| Variable | CPT |
|---|---|
| $i = 1$ | $x_i \succ \bar{x}_i \succ \bar{\bar{x}}_i$ |
| $1 < i \leq k+1$ | $x_{i-1}$ : $x_i \succ \bar{x}_i \succ \bar{\bar{x}}_i$<br>$\bar{x}_{i-1}$ : $\bar{\bar{x}}_i \succ \bar{x}_i \succ x_i$<br>$\bar{\bar{x}}_{i-1}$ : $x_i \succ \bar{x}_i \succ \bar{\bar{x}}_i$ |
| $k+1 < i \leq n$ | $x_{i-1}$ : $\begin{cases} \bar{x}_i \succ x_i \\ \bar{x}_i \succ \bar{\bar{x}}_i \end{cases}$<br>$\bar{\bar{x}}_{i-1}$ : $\begin{cases} x_i \succ \bar{x}_i \\ \bar{\bar{x}}_i \succ \bar{x}_i \end{cases}$ |

Figure 14:

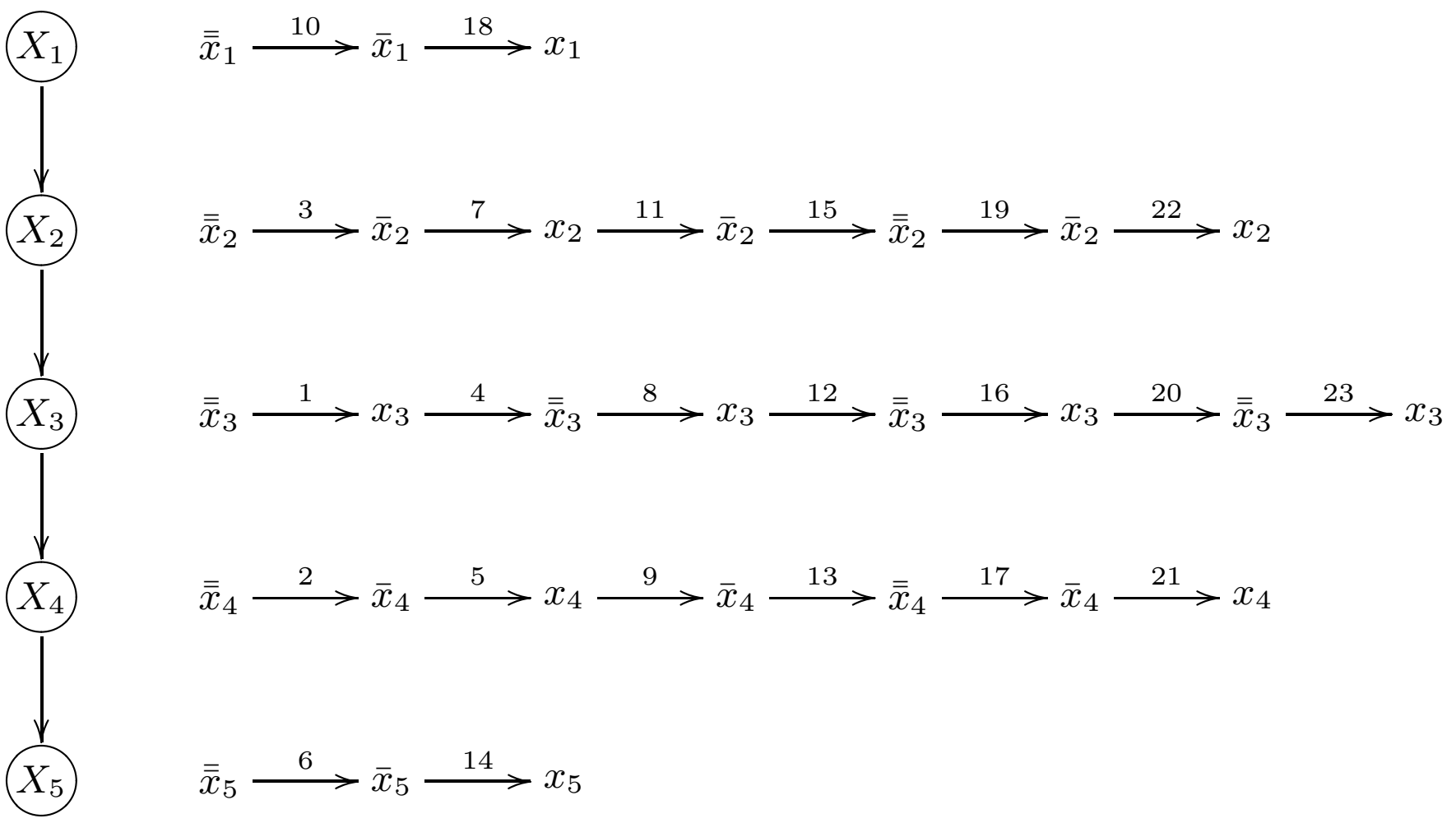

Figure 15:

Now we prove that the size of the minimal flipping sequence for the dominance query as above is greater than $\sum_{i=1}^{k} a_i$. We divide the proof into four steps, where the first step shows the necessity of $\sum_{i=1}^{k} a_i$ flips for any flipping sequence from $o'$ to $o$ as above, and the subsequent three steps prove the existence of a flipping sequence from $o'$ to $o$. Recall that, since $N$ forms a chain, for $1 < i \leq n$, $X_{i-1}$ is the only parent of $X_i$, thus the value flips of $X_i$ depend only on the value of $X_{i-1}$.

The steps of the proof are as follows:

(1) On a flipping sequence from $o'$ to $o$, for $2 \geq i \leq k+1$, every variable $X_{k+i}$ (last $k$ variables) *must* change its value at least $a_{k+2-i}$ times, such that if $f_1^{k+i}, \ldots, f_{a_{k+2-i}}^{k+i}$ is the corresponding sequence of flips of $X_i$, then, for $1 \leq j \leq a_{k+2-i}$, we have:

$$f_j^{k+i} = \begin{cases} \bar{\bar{x}}_i \rightarrow \bar{x}_i, & j = 4k+1 \\ \bar{x}_i \rightarrow x_i, & j = 4k+2 \\ x_i \rightarrow \bar{x}_i, & j = 4k+3 \\ \bar{x}_i \rightarrow \bar{\bar{x}}_i, & j = 4k \end{cases} \qquad k \in \mathbb{N} \tag{7}$$

(2) For $1 \leq i \leq k$, every variable $X_i$ (first $k$ variables) *can* change its value $a_i$ times, such that if $f_1^i, \ldots, f_{a_i}^i$ is the corresponding sequence of flips of $X_i$, then, for $1 \leq j \leq a_i$, every flip $f_j^i$ is consistent with Eq. 7.

(3) Given the sequence of flips for $X_k$ as in (2), the variable $X_{k+1}$ *can* change its value $a_k + 1$ times, such that if $f_1^{k+1}, \ldots, f_{a_k+1}^{k+1}$ is the corresponding sequence of flips of $X_{k+1}$, then, for $1 \leq j \leq a_k + 1$, we have:

$$f_j^{k+1} = \begin{cases} \bar{\bar{x}}_{k+1} \rightarrow x_{k+1}, & j = 2k+1 \\ x_{k+1} \rightarrow \bar{\bar{x}}_{k+1}, & j = 2k \end{cases} \qquad k \in \mathbb{N} \tag{8}$$

(4) The sequence of flips for $X_{k+2}$ as in (1) is feasible given the sequence of flips for $X_{k+1}$ as in (3).

Step (1): The proof is by induction on $i$. For $i = k+1$, the variable $X_{2k+1}$ (i.e., $X_n$) must change its value at least twice, since no value of $X_{2k}$ induces $\bar{\bar{x}}_{2k+1} \prec x_{2k+1}$. The only way to change the value of $X_{2k+1}$ from $\bar{\bar{x}}_{2k+1}$ to $x_{2k+1}$ is to flip $X_{2k+1}$ first from $\bar{\bar{x}}_{2k+1}$ to $\bar{x}_{2k+1}$, and then from $\bar{x}_{2k+1}$ to $x_{2k+1}$. Note that these flips are consistent with Eq. 7 and $a_{k+2-(k+1)} = a_1 = 2$, thus we established the induction base.

Now, we assume that the induction hypothesis that, for $2 < l \leq i \leq k+1$, every variable $X_i$ must change its value at least $a_{k+2-i}$ times according to Eq. 7, and prove this claim for $X_{k+l-1}$.

Consider the sequence of flips of $X_{k+l}$ that we assumed to be necessary, i.e., $f_1^{k+l}, \ldots, f_{a_{k+2-l}}^{k+l}$. According to $CPT(X_{k+l})$ we have that:

(i) every pair of successive flips of $X_{k+l}$ require different values of $X_{k+l-1}$,

(ii) the required values of $X_{k+l-1}$ are $\{\bar{\bar{x}}_{k+l-1}, x_{k+l-1}\}$, and

(iii) the first flip of $X_{k+1}$ (from $\bar{\bar{x}}_{k+l}$ to $\bar{x}_{k+l}$) requires $X_{k+l-1} = x_{k+l-1}$, while $o'[X_{k+l-1}] = \bar{\bar{x}}_{k+l-1}$.

Therefore, $X_{k+l-1}$ must perform $a_{k+2-l}$ value changes from $\bar{\bar{x}}_{k+l-1}$ to $x_{k+l-1}$ and back, in order to support the required value changes of $X_{k+l}$. In addition, after supporting the value changes of $X_{k+l}$, the variable $X_{k+l-1}$ must perform another value change from $\bar{\bar{x}}_{k+l-1}$ to $x_{k+l-1}$, since $a_{k+2-l}$ is even (see Eq. 6), while $o'[X_{k+l-1}] = \bar{\bar{x}}_{k+l-1}$ and $o[X_{k+l-1}] = x_{k+l-1}$.

Finally, according to $CPT(X_{k+l-1})$, every value change of $X_{k+l-1}$ from $\bar{\bar{x}}_{k+l-1}$ to $x_{k+l-1}$ and from $x_{k+l-1}$ to $\bar{\bar{x}}_{k+l-1}$ consists of two flips: from the initial value to $\bar{x}_{k+l-1}$, and from $\bar{x}_{k+l-1}$ to the target value. Thus the proved that $X_{k+l-1}$ must perform at least $2(a_{k+2-l}+1) = a_{k+3-l}$ value changes, and that these value changes are consistent with Eq. 7.

Step (2): The proof is by induction on $i$. For $X_1$ the proof is straightforward, since $o'[X_1] = \bar{\bar{x}}_1$, $o[X_1] = x_1$, and $CPT(X_1)$ allows us to flip the value of $X_1$ from $\bar{\bar{x}}_1$ to $\bar{x}_1$, and then from $\bar{x}_1$ to $x_1$.

Now we assume the induction hypothesis that, for $1 \leq i < k$, every variable $X_i$ can change its value $a_i$ times according to Eq. 7, and prove the claim for $X_{i+1}$. The proof is apparent from $CPT(X_{i+1})$, the outcomes $o$ and $o'$ in query, and the induction hypothesis. For every value achieved by $X_i$ along its sequence of $a_i$ flips (including the initial value $o'[X_i] = \bar{\bar{x}}_i$), $X_{i+1}$ can flip its value twice: Given $X_i = \bar{\bar{x}}_i$ or $X_i = x_i$, we can flip the value of $X_{i+1}$ first from $\bar{\bar{x}}_{i+1}$ to $\bar{x}_{i+1}$, and then from $\bar{x}_{i+1}$ to $x_{i+1}$. Alternatively, given $X_i = \bar{x}_i$, we can flip the value of $X_{i+1}$ from $x_{i+1}$ to $\bar{x}_{i+1}$, and then from $\bar{x}_{i+1}$ to $\bar{\bar{x}}_{i+1}$. Therefore, $X_{i+1}$ can flip its value $2(a_i + 1) = a_{i+1}$ times, and it is easy to see that these value flips of $X_{i+1}$ are consistent with Eq. 7.

Step (3): Given the sequence of $a_k$ flips $f^k_1, \ldots, f^k_{a_k}$ of $X_k$ as in Eq. 7, let $\mathsf{val}(f^k_j) \in Dom(X_k)$ be the value of $X_k$ achieved by the flip $f^k_j$, for $1 \leq j \leq a_k$. $CPT(X_{k+1})$ entails that, for any triple of values of $X_k$ achieved by a triple of successive flips $f^k_j, f^k_{j+1}, f^k_{j+2}$, $1 \leq j \leq a_k - 2$, we have either:

$$\begin{array}{rcl} \mathsf{val}(f^k_j) & : & x_{k+1} \succ \bar{\bar{x}}_{k+1} \\ \mathsf{val}(f^k_{j+1}) & : & \bar{\bar{x}}_{k+1} \succ x_{k+1} \\ \mathsf{val}(f^k_{j+2}) & : & x_{k+1} \succ \bar{\bar{x}}_{k+1} \end{array}$$

or

$$\begin{array}{rcl} \mathsf{val}(f^k_j) & : & \bar{\bar{x}}_{k+1} \succ x_{k+1} \\ \mathsf{val}(f^k_{j+1}) & : & x_{k+1} \succ \bar{\bar{x}}_{k+1} \\ \mathsf{val}(f^k_{j+2}) & : & \bar{\bar{x}}_{k+1} \succ x_{k+1} \end{array}$$

In addition, we know that:

$$o'[X_k] = \bar{\bar{x}}_k \;\; : \;\; x_{k+1} \succ \bar{\bar{x}}_{k+1}$$

and

$$\mathsf{val}(f^k_1) = \bar{x}_k \;\; : \;\; \bar{\bar{x}}_{k+1} \succ x_{k+1}$$

The above entails that $X_{k+1}$ can change its value $a_k + 1$ times according to Eq. 8.

Step (4): The proof of this subclaim is straightforward from $CPT(X_{k+2})$. Given the sequence of $X_{k+2}$ value flips $f^{k+2}_1, \ldots, f^{k+2}_{a_k}$ as in Eq. 7, observe that, for $1 \leq j \leq a_k$, if $j = 2k + 1$ for some $k \in \mathbb{N}$, then $f_j^{k+2}$ can be supported by the value $x_{k+1}$ of $X_{k+1}$, otherwise, if $j = 2k$, $k \in \mathbb{N}$, then $f_j^{k+2}$ can be supported by the value $\bar{\bar{x}}_{k+1}$ of $X_{k+1}$. Now, since for $1 \leq l \leq a_k + 1$,

$$\mathsf{val}(f_l^{k+1}) = \begin{cases} x_{k+1}, & l = 2m + 1 \\ \bar{\bar{x}}_{k+1}, & l = 2m \end{cases} \qquad m \in \mathbb{N}$$

it is apparent that the sequence of $a_k$ value flips of $X_{k+2}$ as in Eq. 7 is feasible given the sequence of $a_k + 1$ value flips[12] of $X_{k+1}$ as in Eq. 8.

The above entails that $N \models o \succ o'$, and that the size of the minimal flipping sequence from $o'$ to $o$ in $N$ is $\sum_{i=1}^{k} a_i$, which is greater than $2^{\frac{n}{2}}$. In fact, it can be show that the value flips for $X_1, \ldots, X_{k+1}$, constructed in steps (2)-(3), are the part of the minimal flipping sequence from $o'$ to $o$, thus the length of this sequence is greater than:

$$2 \sum_{i=1}^{k} a_i \ + \ (a_k + 1)$$

However, the result achieved in step (1) already proves our claim that there exist dominance queries on multi-valued CP-nets with partially specified preferences, for which there are exponentially sized minimal flipping sequences. $\square$

12. In fact, only $a_k$ value flips of $X_{k+1}$ are used in order to perform $a_k$ value flips of $X_{k+2}$ as required. The last flip of $X_{k+1}$ is used in order to achieve the value $o[X_{k+1}]$ for $X_{k+1}$ after supporting $X_{k+2}$.